\documentclass[final,3p,times]{elsarticle}
\usepackage[dvipsnames,svgnames,x11names]{xcolor}
\usepackage[draft]{changes}
\usepackage[inline]{enumitem}
\definechangesauthor[color=blue]{PC}
\usepackage{lineno,hyperref}
\usepackage{amsmath,amssymb,amsfonts}
\usepackage{algorithmic}
\usepackage{graphicx}
\usepackage{anyfontsize}
\usepackage{adjustbox}
\usepackage{rotating}
\usepackage{tablefootnote}
\usepackage{threeparttable}
\usepackage{longtable}
\usepackage{tabularray}
\usepackage{caption}
\usepackage{subfigure}
\usepackage[font=small,labelfont=bf, figurename=Fig.]{caption} 

\usepackage{textcomp}
\usepackage{todonotes}
\usepackage{tabularx}
\usepackage{gensymb}
\usepackage[utf8]{inputenc}
\usepackage{mathtools, nccmath}
\usepackage{booktabs}
\usepackage{cleveref}
\usepackage{chemformula}
\usepackage[version=4]{mhchem}
\usepackage{booktabs, multirow} 
\usepackage{soul}
\modulolinenumbers[1]

\biboptions{authoryear}

\usepackage{comment}
\usepackage[inline]{enumitem}   
\makeatletter
\newcommand{\inlineitem}[1][]{%
\ifnum\enit@type=\tw@
    {\descriptionlabel{#1}}
  \hspace{\labelsep}%
\else
  \ifnum\enit@type=\z@
       \refstepcounter{\@listctr}\fi
    \quad\@itemlabel
    \hspace{\labelsep}%
\fi}
\makeatother
\begin{document}

\begin{frontmatter}

\title{A Review of Driver Gaze Estimation and Application in Gaze Behavior Understanding}



\author[mymainaddress]{Pavan Kumar Sharma}

\author[mymainaddress]{Pranamesh Chakraborty\corref{mycorrespondingauthor}}

\cortext[mycorrespondingauthor]{Corresponding author\\
Pavan Kumar Sharma: pavans20@iitk.ac.in, Pranamesh Chakraborty: pranames@iitk.ac.in,+91-512-259-2146}

\address[mymainaddress]{Department of Civil Engineering, Indian Institute of Technology Kanpur, Kanpur-208016, U.P, India}

\begin{abstract}
\par Driver gaze plays a key role in different gaze-based applications, such as driver attentiveness detection, visual distraction detection, gaze behavior understanding, and building driver assistance system. The main objective of this study is to perform a comprehensive summary of driver gaze fundamentals, methods to estimate driver gaze using machine learning (ML) based technique, and its applications in real world driving scenarios. We first discuss the fundamentals related to driver gaze, involving head-mounted and remote setup based gaze estimation and the terminologies used for driver gaze behavior understanding. Next, we list out the existing benchmark driver gaze datasets, highlighting the collection methodology and the equipment used for such data collection. This is followed by a discussion of the algorithms used for driver gaze estimation, which primarily involves traditional machine learning and deep learning (DL) based techniques. The estimated driver gaze is then used for understanding gaze behavior while maneuvering through intersections, on-ramps, off-ramps, lane changing, determining the effect of roadside advertising structures and also for developing driver gaze based applications such as maneuver prediction, driver inattention, and distraction detection systems, etc. Finally, we discuss the limitations in the existing literature, challenges, and future scope in driver gaze estimation and gaze-based applications.

\end{abstract}
\begin{keyword}
Driver gaze \sep  gaze estimation \sep driver gaze datasets \sep driver gaze behavior understanding 
\end{keyword}

\end{frontmatter}
\section{Introduction}
Driver safety is one of the major global concerns due to the increasing number of road crashes yearly. According to the Global status report on road safety 2018 by World Health Organization (WHO), approximately 1.3 million people die every year from road crashes \citep{world2018global}. There are several causes of road crashes, out of which distracted driving, drowsiness, and inattentiveness of the driver from their surrounding traffic are also significant. Driver gaze is an important clue to measuring distraction \citep{koay2022detecting, misra2023detection} and attentiveness \citep{jha2023driver} to the surroundings. Although research on driver gaze estimation has been mostly carried out during the last two to three decades, however, the history of human gaze estimation dates back to the nineteenth century. In the early 20th century, it was limited to the medical field with invasive gaze-estimation techniques \citep{mowrer1935corneo}. Due to the advancement of technology over the past few decades, gaze estimation has become one of the critical research field. In addition to driver gaze estimation, estimation of human gaze is used in many other applications, such as human-computer interaction \citep{majaranta2014eye, shimata2015study, prabhakar2021brief, pathirana2022eye}, health care and medical field \citep{hansen2009eye, konig2014nonparametric, harezlak2018application}, education and e-learning \citep{rosch2013review, sun2017application},  consumer psychology and marketing \citep{recarte2008mental,tomas2021goo, senarath2022customer, pathirana2022single}, etc. In the driving context, the gaze can be estimated using intrusive or non-intrusive methods. In an intrusive technique, drivers wear a head-mountable eyeglass setup or eye trackers, some of which look like normal eyeglasses. In contrast, non-intrusive techniques, remote gaze tracking systems are used, which classify the driver gaze in several areas of interest (AOI). In the literature, AOI is also known as gaze zone or gaze class. In this paper, we will primarily use the term gaze zone in the case of remote gaze tracking.
{\color{black}\par Several review papers exist on gaze tracking \citep{kar2017review, khan2019gaze, akinyelu2020convolutional, klaib2021eye, shehu2021remote, pathirana2022eye, wang2021vision, zheng2022opportunities} in different fields such as manufacturing and logistics, consumer psychology and marketing, human-computer interaction (HCI) etc. A few of these studies also focus on driver gaze tracking \citep{kar2017review, klaib2021eye, pathirana2022eye}. However, these studies have not done an in-depth analysis of driver gaze tracking and did not discuss how it differs from gaze tracking in other platforms, specifically in terms of metrics, feature extraction, and algorithms used for driver gaze tracking.  {\color{black}\citet{prabhakar2021brief} reviewed different interactive automotive user interfaces, including the gaze controlled ones which can help drivers perform secondary tasks while driving. However, this study didn't cover driver gaze behavior understanding and other driver gaze related application areas.}

\par A few studies have focused on driver gaze tracking and its application in inattention, distraction, and drowsiness detection.
\cite{koay2022detecting}  primarily focused on different models of driver inattention measurement for assistive and autonomous driving and discussed driver face and eye datasets. However, this study focused on driver distraction detection using various physiological (e.g., EEG, ECG) and external sensors (steering angle, GPS, etc.) along with visual sensors for gaze tracking (camera types and eye tracker). On the other hand, \citet{khan2019gaze} and \citet{kotseruba2022attention} reviewed different driver eye-tracking techniques and attention models, specifically related to driver gaze. However, the discussion was limited to different inattention, distraction, and drowsiness detection techniques, which can be used to measure attention levels and build driver monitoring systems and advanced driver assistance systems (ADAS).  They did not discuss regarding the uses of driver gaze tracking in understanding driver behavior in complex situations such as maneuvering through intersection, lane-changing and overtaking behavior and its significance for safer road infrastructure design. To fill these existing gaps, the present review study primarily makes the following contributions:}

\par
\begin{enumerate}[itemsep=0pt,parsep=0pt,topsep=0pt,partopsep=0pt]
    \item Discussion on metrics for driver gaze used in both head-mounted and remote setup and their relationship to each other.
    \item In-depth analysis of the features and models specifically useful for driver gaze estimation.
    \item Uses of driver gaze specifically in driver behavior understanding along with application in different driving fields such as building driver maneuver recognition and prediction systems, ADAS along with distraction detection and inattention measurement system.
    \item Discuss different driver gaze estimation challenges and their possible solutions and future scope.
\end{enumerate}

\begin{figure}[!htb]
    \centering
   \includegraphics[width=0.8\textwidth]{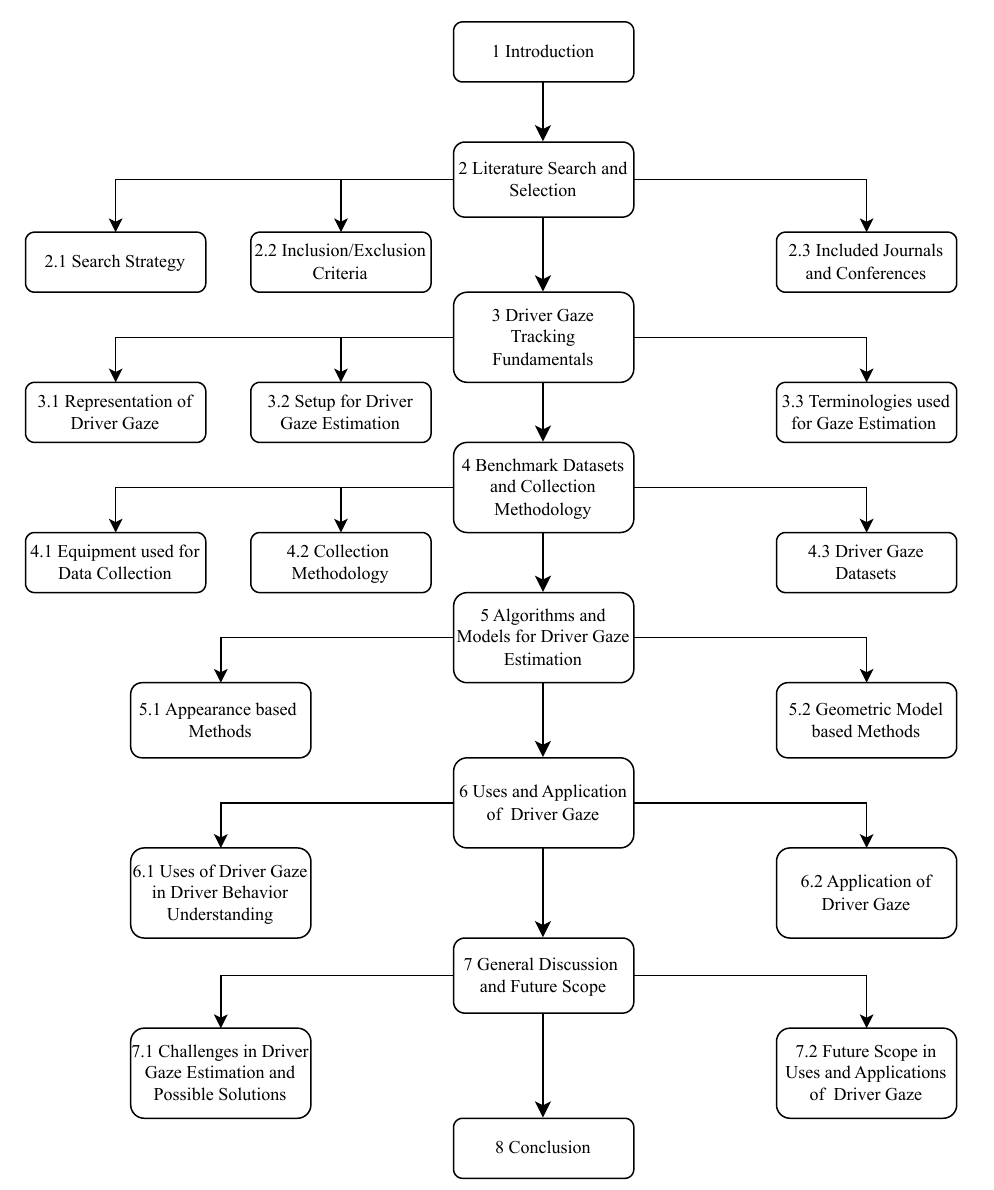}
    \caption{Structure of the paper}
    \label{fig:structure of the paper}
\end{figure}

The structure of the paper is illustrated in Fig. \ref{fig:structure of the paper} and the remainder of the paper is organized as follows. Section 2 describes the literature search and selection criteria for review, followed by Section 3, which describes driver gaze representation,  driver gaze estimation setups, and different terminologies related to driver gaze.
Existing driver gaze benchmark datasets with their pros and cons are covered in Section 4, while different gaze estimation algorithms and models used in appearance-based and geometric model-based methods are discussed in Section 5. Section 6 covers the uses of driver gaze in gaze behavior understanding and different driver gaze-based applications, such as building advanced driving assistance systems. Finally, Section 7 provides a general discussion and future scope, including the existing studies challenges and limitations, followed by a  conclusion at the end of the paper.

\section{Literature Search and Selection}

\begin{figure}[!htb]
    \centering
   \includegraphics[width=0.8\textwidth]{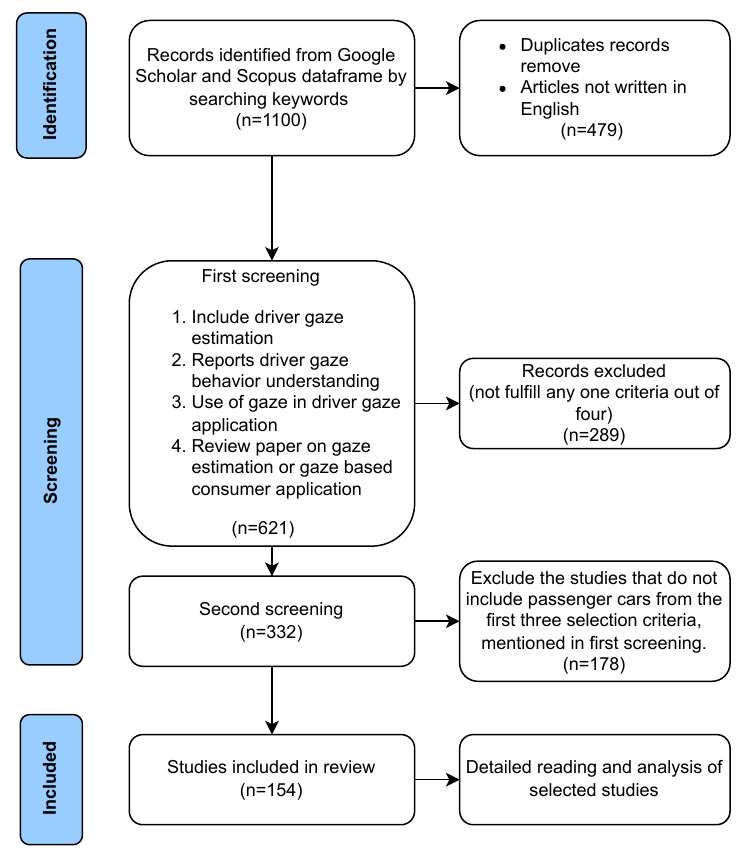}
     \caption{Literature selection criteria based on PRISMA technique}
    \label{fig:Literature Selection Criteria}
\end{figure}

In this review study, we have conducted a systematic literature review using the Preferred Reporting Items for Systematic reviews and Meta-Analyses (PRISMA) 2020 report \citep{page2021prisma}. Following the PRISMA guidelines, we have followed the search strategy, exclusion and inclusion criteria (see Fig. \ref{fig:Literature Selection Criteria}), details of which are discussed next.

\subsection {Search strategy}
We have passed a thorough search on Google Scholar and Scopus using search keywords, which can be categorized into three categories. The first category is related to gaze estimation, predominantly for driver gaze. It includes the keywords: \emph{gaze estimation, driver gaze estimation, driver head and eye pose estimation, gaze tracking,  driver gaze, and eye tracking}. The second category is related to driver gaze-based behavior understanding, which includes the keywords:  \emph{driver gaze behavior, driver gaze behavior at intersections, lane changing gaze behavior, and over passing gaze behavior}. The third category is related to diver gaze-based applications, which includes the keywords: \emph{driver gaze distraction, driver inattentiveness, gaze-based advanced driving assistance system, and gaze-based driver warning system.}

\subsection{Inclusion/Exclusion criteria}

We have considered the papers which are published in scientific journals and conference proceedings. Initially, literature samples included approximately 1100 papers from Google Scholar and Scopus using the search keywords discussed above. After removing the duplicates, 621 remaining papers were considered for screening (second stage). In the first stage of screening, the remaining papers were screened based on the papers' titles, abstract, keywords and dissemination outlets. We have considered the following selection criteria for this stage. 1) selected paper includes driver gaze estimation, 2) reports results of driver gaze behavior understanding, 3) use of driver gaze in any application, such as driver assistance, driver safety, warning or alert system, 4) review papers on gaze estimation or gaze application in different consumer platforms such as retail stores, manufacturing and logistics, navigate menus in smart TVs, gaze-based input keyboards for laptops, etc. Out of the above four criteria, if anyone is satisfied, the paper is considered for the second screening stage. In second screening stage, selected papers that fulfill first three criteria are further checked to find whether they have considered passenger car or not. Only the passenger car-based study has been considered for the refined list of paper analyses. Finally, 154 relevant research papers published until June 30, 2023, have been considered in this present study.

\subsection{Included journals and conferences}
 The research papers reviewed in this study are from different reputed journals such as Knowledge-Based Systems, Expert Systems with Applications, Engineering Applications of Artificial Intelligence, Transportation Research Part A: Policy and Practice, Transportation Research Part C: Emerging Technology, Transportation Research Part F: Traffic Psychology and Behavior, Transportation Research Interdisciplinary Perspectives, Transportation Engineering (TRENG), Journal of Eye Movement Research, International Journal of Human–Computer Interaction (IJHCI), Journal of Safety Research, Pattern Analysis and Machine Intelligence, Transactions on Intelligent Transportation Systems, Transactions on Intelligent Vehicles, Transactions on Vehicular Technology, Transactions on Information Theory, IET Intelligent Transport Systems, Spanish Journal of Psychology, International Journal of Robotics Research, etc. We have also included conference papers such as Intelligent Transportation Systems Conference, Intelligent Vehicles Symposium (IV), Computer Vision and Pattern Recognition (CVPR), International Conference on Computer Vision (ICCV), European Conference on Computer Vision (ECCV),  International Conference on Automatic Face and Gesture Recognition, International Conference on Robotics and Biomimetics, International Conference on Learning representations (ICLR), Association for Computing Machinery (ACM),  International Conference on Automotive User Interfaces
and Interactive Vehicular Applications (AutoUI), International Conference on Intelligent User Interfaces (IUI), and Conference on Human Factors in Computing Systems (CHI). 

\section{Driver Gaze Tracking Fundamentals}
In the literature, estimation of driver gaze is referred by different terminologies, such as eye pose estimation, eye tracking  \citep{liu2002real}, gaze estimation, gaze detection \citep{naqvi2018deep, yoon2019driver, rangesh2020driver}, and gaze tracking. However, all these are similar and often used interchangeably in different research papers. This section begins with a discussion of how driver gaze output is represented in the driving field. Subsequently, head-mounted and remote setups for driver gaze estimation are explored, highlighting their advantages and disadvantages. The discussion concludes with an overview of key terminologies used for driver gaze behavior understanding and gaze-based different driving applications.

\subsection{Representation of driver gaze}
\begin{figure}[!htbp]
    \centering
   \includegraphics[width=0.5\textwidth]{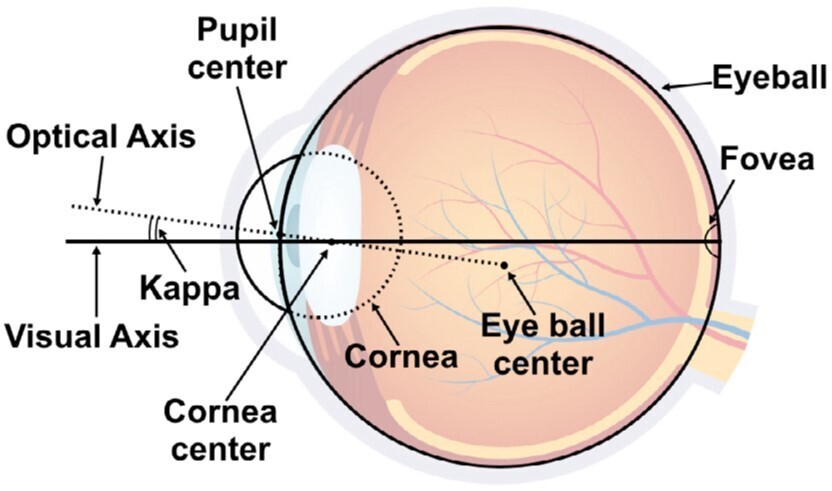}
     \caption{Geometric model of human eye \citep{pathirana2022eye}}
    \label{fig:3D eye model}
\end{figure}

In the driving field, gaze estimation is a technique to determine where a driver is looking. 
Existing literature on driver gaze provides insight into three common methods of gaze measurement. Firstly, the zone-based approach categorizes the driver gaze location into different discrete gaze zones. These zones encompass areas such as the windshield, dashboard, forward, speedometer, rearview mirror, left-wing mirror, right-wing mirror, etc. The output of this method is the gaze zone class, offering a discrete type of driver gaze estimation \citep{chuang2014estimating, tawari2014attention, diaz2016reduced, vora2018driver, ribeiro2019driver, ortega2020dmd, ghosh2021speak2label}. The second approach of driver gaze measurement is described using the gaze direction, gaze vector, and point of gaze. The gaze direction is defined as the ray that goes through the cornea centre and the fovea. 
{\color{black} As shown in Fig. \ref{fig:3D eye model}, the deviation (Kappa) of the visual axis from the optical axis determines the gaze direction.} The point on the plane or the object surface on which the gaze direction intersects is called the point of gaze \citep{sun2016auxiliary}. In a head-mounted setup, the point of gaze is the intersecting point on the scene (object's surface) captured by the scene camera. In contrast, a gaze vector is the mathematical representation of gaze direction in 3D space \citep{yang2019dual, yuan2022self}. It typically includes three components: (x, y, z), where each component represents the gaze direction along one of the coordinate axis, where x and y represent gaze in horizontal and vertical directions while z represents the depth component of gaze. The depth component z provides information about how far or near the driver is looking.
The third approach of gaze measurement is based on the gaze object or entity they are looking at, such as a car, bus, truck, traffic light, or pedestrian. In this case, the output of driver gaze estimation is the class of the gaze traffic entity \citep{palazzi2018predicting, dua2020dgaze}.  Traditionally, gaze zone-based driver gaze estimation has been mostly prevalent in the literature. However, a recent shift in trend indicates a preference for measuring driver gaze in terms of the gaze direction or point of gaze, which is a more refined and precise method of gaze estimation.

\subsection{Setup  for driver gaze estimation}
Gaze estimation is done using different techniques depending on the application. Initially, sensors attached to the facial skin, such as electrode pairs were used to record potential differences during eye movements \citep{mowrer1935corneo} to understand cognitive behavior. This technique, though accurate, is usually uncomfortable to the users. Due to the advancement of computer vision-based technology, gaze estimation has seen widespread application in different fields, including driver assistance and behavior understanding. Head-mounted  (wearable) and remote setup (non-wearable) are the two commonly used  setups in practice for driver gaze estimation, which are briefly discussed next.

\subsubsection{Head mounted gaze estimation}
\begin{figure}
    \centering
   \includegraphics[width=0.9\textwidth]{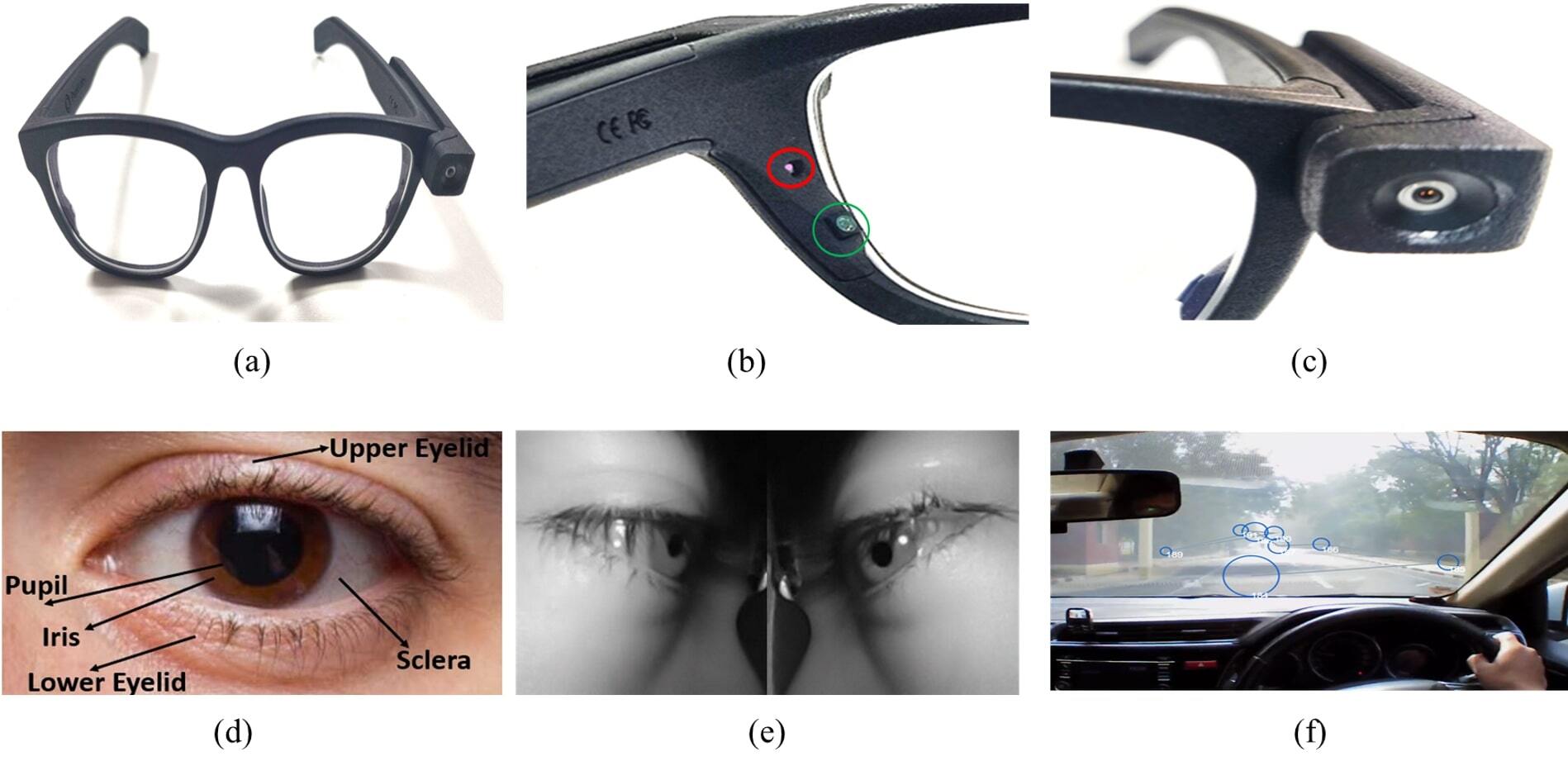}
    \caption{(a) Eye tracking glass (Pupil Invisible) (b)  Near Eye camera (green circle) and LED light (red circle) (c) Scene camera (d) Different parts of an Eye (e) Left and Right Eye Infrared images captured by using Near Eye Cameras (f) Driver frontal view recorded by Scene Camera (blue circles represent fixations).}
    \label{fig:Pupil invisible Eye glasses and its parts}
\end{figure}

In head-mounted gaze estimation, driver mounts or wear a device on their head called an eye tracker. Fig. \ref{fig:Pupil invisible Eye glasses and its parts}a shows a sample eye-tracker glass, which looks exactly similar to regular prescription glasses. Primarily head-mounted system consists of near-eye cameras and infrared LED light for active illumination of the eyes (Fig. \ref{fig:Pupil invisible Eye glasses and its parts}b) and a scene camera (Fig. \ref{fig:Pupil invisible Eye glasses and its parts}c). In this system, a near-eye camera is present near each eye, recording the eye movements from close-up, as shown in Fig. \ref{fig:Pupil invisible Eye glasses and its parts}e. The scene camera records the frontal view, allowing correlation of gaze data to the cues and stimuli present in the scene of the driver (Fig. \ref{fig:Pupil invisible Eye glasses and its parts}f).

Head-mounted eye trackers rely on detecting image features for the near eye cameras, including the pupil, iris contours (Fig. \ref{fig:Pupil invisible Eye glasses and its parts}d),  and glints, i.e., reflections produced by the Infrared LEDs.  Typically, eye trackers require calibration of each driver before estimating the driver gaze \citep{dukic2012older, scott2013visual, lemonnier2015gaze, lemonnier2020drivers}. However, some latest eye trackers have been developed which are calibration-free \citep{tonsen2020high}. 
Several studies have used eye trackers to understand the driver gaze behavior in indoor (simulation-based studies) \citep{romoser2013comparing, scott2013visual, lemonnier2015gaze} and outdoor traffic environments (real-world driving studies) \citep{bao2009age, dukic2012older, li2019drivers, lemonnier2020drivers}. 
In these studies, driver gaze output, has been measured in any of the three forms (gaze zones, gaze direction, gaze on traffic entity), which are discussed in Section 3.1.
\par Head-mounted gaze estimation in the driving field offers significant insights into driver gaze behavior understanding and gaze-based application.  
The primary advantages of this gaze estimation setup are high accuracy, fine-grained gaze data, and reduced environmental noise. Since the driver wears the head-mounted gaze estimation device (see Fig. \ref{fig: Gaze zone estimation techniques}a), it captures the eyes and head movement information from a very close distance. Because of this, the gaze estimation accuracy is relatively high, and the output is typically gaze direction or point of gaze. This system can capture gaze-related metrics such as fixation, saccade, and pupil dilation of driver gaze, which will be discussed next. This setup is also relatively less affected by environmental factors such as adverse lighting conditions. However, the head-mounted gaze estimation system has some common disadvantages, such as intrusiveness, limited comfort and accessibility, and frequent calibration. Head-mounted gaze estimation systems typically require the driver to wear a device on their head, which can be uncomfortable and intrusive, especially for those who do not wear regular eyeglasses. It is challenging to continuously use the system for long drives. This can lead to discomfort, distraction, or reduced situational awareness, which is counterproductive for safe driving. Further proper calibration is crucial for the accuracy of head-mounted gaze estimation systems. Calibrating the system correctly for each user can be time-consuming, and inaccuracies in calibration can lead to misleading results.

\subsubsection{Remote setup gaze estimation}
For remote setup-based gaze estimation, cameras are typically placed on the dashboard \citep{chuang2014estimating, vicente2015driver, naqvi2018deep, yoon2019driver, wang2019continuous} or sometimes mounted on the windshield \citep{choi2016real}. Single or multiple cameras can be used, depending on the nature of gaze zone classification. Typically, single cameras are preferred when dashboard and windshield areas are divided into coarser gaze zones, while in finer gaze zones classification, multiple cameras are primarily used \citep{naqvi2018deep, yoon2019driver, guo2019generalized}. A detailed discussion on coarser and finer gaze zones is given in Section 4.2. Multiple cameras capture face images from different angles, such as left and right sides of the face and eye \citep{yoon2019driver}. This helps to make gaze estimation robust even for substantial head movement, which is a shortcoming of single camera-based estimation. Typically, the output of remote setup gaze estimation is the gaze zone class, but in very few studies \citep{yang2019dual,yuan2022self}, the output is the point of gaze. 

The non-intrusive nature of the remote gaze estimation system eliminates some of the limitations of head-mounted gaze estimation. The driver is not required to wear any head-mounted and intrusive devices, which can lead to a more natural and comfortable driving experience. However, these systems typically have lower accuracy compared to the head-mounted gaze estimation technique. Since they are placed on the dashboard or mounted on a windshield, it has difficulty capturing detailed information about eye movements, such as the position or movement of the pupil and cornea inside the eye, making it less accurate than head-mounted device. {\color{black}Most remote setup systems} do not capture gaze metrics data such as fixation and pupil dilation, which is essential for different driver applications. Further, variations in driver posture and seat position also affect the gaze estimation accuracy.

\subsection{Terminologies used for gaze estimation}
This section discusses the different terminologies related to head-mounted and remote setup based temporal characteristics of driver gaze. 
This involves studying the duration for which a driver looks or focuses his/her attention on a particular object or region. Typically, the terms fixation, saccades, scanpath, blink rate, and pupil dilation are related to temporal gaze estimation in head-mounted setup, where the driver gaze is estimated as the point of gaze. On the other hand, in remote setup based gaze estimation, it is challenging to capture the exact eye movement. Therefore, instead of the point of gaze, gaze zone class is the estimated gaze output. In this case, primarily, the term glance is related to driver gaze, which depends on the driver's head pose and eye pose.

\begin{enumerate}
    \item \label{item:1} Fixation: It is the state of eyes in which a driver maintains the visual gaze in an area of interest (AOI) for a certain period \citep{dukic2012older, kar2017review}.  
    Most fixations last between 100 to 600 milliseconds, but this varies based on the context and other relevant factors \citep{pauszek2023introduction}. Remote eye tracker manufacturers such as Smart Eye Pro and Tobii consider 200 ms as the minimum gaze period for a valid fixation in driver behavior studies \citep{holmqvist2022eye}. The following terms further describe fixation.
        \begin{enumerate}
            \item \label{subitem:1a} Dwell Time: It is the sum of the duration of fixations in a given area of interest. In driver behavior studies, dwell time gives the proportion of time spent by a driver gazing at an object in an interval of time. The object can be dynamic objects such as moving vehicles or static objects such as stationary vehicles, traffic lights, traffic signs, road markings, etc. \citep{lemonnier2015gaze, chung2022static}. Higher dwell time in a given gaze zone represents a high level of interest of the driver in that particular zone.
            \item \label{subitem:1b}Time to first fixation (TTFF): It measures how long a driver takes to start looking at a specific stimulus in a gaze zone after the stimulus is presented. TTFF indicates how much an aspect of the scene initially attracted attention.
            \item \label{subitem:1c} Number of fixations: It represents the total number of times a subject fixates their gaze in a given gaze zone in a given time interval \citep{chung2022static}.
        \end{enumerate}
    \item \label{item:2} Saccades: Saccades are defined as the rapid eye movement when the gaze sight shifts from one fixation point to another \citep{imaoka2020assessing}. The time taken to shift the eyes from one fixation point to another is called saccadic duration. The majority of saccades range in duration from 20 to 80 milliseconds, although this varies considerably depending on the situation \citep{pauszek2023introduction}. It is also measured using other parameters such as saccade number, amplitude, and fixation-saccade ratio \citep{kar2017review}. It shows the dynamics of the driver gaze while driving.
    \item \label{item:3}Scanpath: Scanpath represents the path followed by the driver's eyes to reach from one area of interest to another. It is also defined as the sequence of fixations and saccades that an individual driver makes while performing a task, such as lane changing, turning at an intersection, etc.   Scanpath includes scanpath direction, duration, and length  \citep{kar2017review, pauszek2023introduction}. 
    Scanpath duration refers to the total time taken to complete a scanpath and it can be measured by summing the cumulative time spent on fixations and saccades during the task. Scanpath length typically refers to the spatial extent covered by the eyes during a scanpath. It represents the total distance travelled by the eyes while performing a task and is often measured in visual degrees or pixels.
    \item \label{item:4} Glances: Glances are the coarser measurement of gaze, while fixation is a more refined measurement. It measures the gaze over an area, while fixation is measured at a point. In an interested gaze zone, glances may contain fixation and saccades. Here, gaze zones are pre-defined areas inside the vehicle cabin, such as the speedometer, center stack, left wing mirror, right wing mirror, rearview mirror \citep{martin2018dynamics, rangesh2020driver, schindler2021truck}, etc. Glances are measured using glance duration, glance frequency, glance transition, glance transition sequence, glance transition length, and number of glance transitions, which are discussed next.
        \begin{enumerate}
            \item \label{subitem:4a} Glance Duration: It is the time spent in each gaze zone in a given time interval. It is similar to the dwell time used to define fixations. It is typically measured in terms of minimum, maximum, and average glance duration made by drivers in different gaze zones \citep{birrell2014glance, martin2018dynamics} to understand driver gaze behavior. Longer glances off the road can be used to detect driver distraction.
            \item \label{subitem:4b} Glance Frequency: The number of glances a driver makes in a given gaze zone in a unit time interval is glance frequency \citep{birrell2014glance}. Several studies include glance duration and frequency in examining the driver attention level while driving \citep{munoz2016distinguishing}. Longer glance duration and higher glance frequency signify the higher task demand \citep{wang2014sensitivity}.
            \item \label{subitem:4c} Glance Transition: It denotes shifting of gaze from one gaze zone to another while assessing the situation of the surrounding traffic environment \citep{scott2013visual, lemonnier2015gaze}. Glance transitions reveal the flow of attention between different gaze zones while driving. A higher correlation exists between the two gaze zone when glance transitions occurring between the two are more frequent \citep{fridman2017can}.
            \item \label{subitem:4d} Glance Transition Sequence: This is the sequence followed by the glances shifting from one gaze zone to another in a given time duration. The occurrence of unusual sequences of gaze patterns while driving contains more rich information than usual. 
            \item \label{subitem:4e} Glance Transition Length: This is defined as the time duration for shifting glances from one specific gaze zone to another. It depends on the position of the former and later gaze zone class. Higher glance transition length can be observed when former and later classes are forward and leftwing mirror, as compared to forward and right wing mirror (see Fig. \ref{fig:Gaze zone classification}a).
            \item \label{subitem:4f} Number of Glance Transition: It is the sum of glance transitions the driver makes from one gaze zone to another in a given time interval. A higher number of glance transitions typically indicates higher driver attentiveness to their traffic surrounding. 
        \end{enumerate}  
    \item \label{item:5} Pupil size (dilation/constriction) and blink rate: Pupil dilation refers to an increase in pupil size, while constriction denotes a decrease in size. In a real-world driving study \citep{heger1998driving}, the mental workload of the driver has been measured based on the blink rate. The findings of this study revealed that as the road curvature increased, the eye blink rate reduced. Blink rate has been found to decrease for visually demanding tasks \citep{marquart2015review}, while shorter blink duration has been observed for mentally and visually demanding tasks \citep{kramer2020physiological}. 
    \item \label{item:6} Entropy Rate: Entropy rate, inspired by the concept of information entropy, is one of the critical metrics used in head-mounted and remote setup-based gaze estimation to measure driver attentiveness to the surrounding traffic.  \citep{shannon1948mathematical}. In a driving scenario, stationary gaze entropy (SGE) and gaze transition entropy (GTE) are two commonly used metrics for measuring driver attention level. 
\par Stationary gaze entropy \citep{bao2009age, chung2022static} describes the information generated by the driver gaze dispersed across the gaze zones. Stationary gaze transition is defined as:

\begin{equation}
S G E=-\sum_{l=1}^L p_{l} \log _2 p_{l}
\end{equation}
where $L$ is the total number of gaze zones. The probability of the driver looking towards gaze zone $l$ is represented by $p_{l}$. SGE does not reflect how drivers control and assess the situation from the surrounding traffic environment. To overcome this limitation, gaze transition entropy \citep{krejtz2015gaze, chung2022static} is used to measure the complexity of the different gaze transition patterns. For instance, when there are multiple stimuli, such as intersections, the gaze pattern is complex compared to an ordinary road; hence the GTE is higher at intersections. 

\begin{equation}
\text { GTE }=-\sum_{k=1}^L p_{k} \sum_{k, l=1}^L\left(p_{k, l}\right) \log _2\left(p_{k, l}\right) \quad\left\{\begin{array}{c}
k, l=1,2,3 \ldots L \\
k \neq l
\end{array}\right.
\end{equation}

Where $p_{k, l}$ is the occurrence of transition probability from gaze zone $k$ to gaze zone $l$.
\end{enumerate}

 \begin{figure}
   \centering
     \includegraphics[width= 0.9\textwidth]{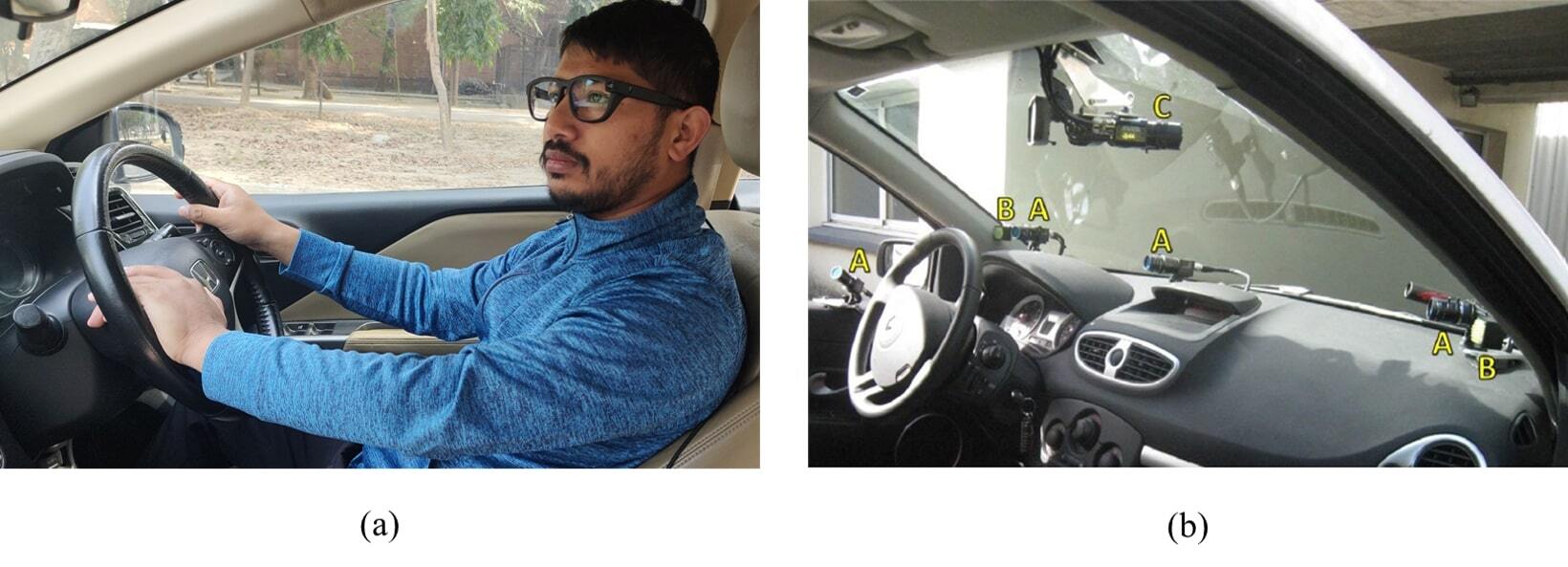}
    \caption{(a) Driver gaze estimation using wearable eye tracker: Driver wearing eye tracking glasses (b) Photograph of instrumented vehicle for remote setup of gaze estimation A:Capture gaze information, B:Infra-red lamp, C:Scene camera \citep{lemonnier2020drivers}.}
     \label{fig: Gaze zone estimation techniques}
 \end{figure}

\section{Benchmark Datasets and Collection Methodology}
Good quality data is one of the crucial needs for computer vision-based gaze-tracking applications. The quality of driver gaze datasets depends on the precision and configuration of collection equipment, methodology, and information level. Driver data must include possible driving scenarios and conditions and a sufficiently large number of subjects. This section discusses the different types of equipment used for data collection, methodology adopted, and different open-source benchmark driver gaze datasets available for gaze estimation model development.

\subsection{Equipment used for data collection}
Driver face data is generally collected using a remote setup, where cameras are installed in front of the driver (on the dashboard or windshield) to record the driver's head movement. Different types of cameras are used to collect driver face data, including traditional RGB (red, green, and blue), RGB-Depth, and Infrared cameras. 
Traditional RGB cameras are used to capture driver face in the visible wavelength spectrum \citep{chuang2014estimating, tawari2014attention, diaz2016reduced, vora2018driver, ribeiro2019driver, ortega2020dmd, ghosh2021speak2label}. However, the image quality sufficiently degrades in low-light conditions, making it difficult to understand driver gaze at night. To alleviate this problem, Infrared (IR)\citep{ribeiro2019driver, rangesh2020driver, ortega2020dmd}/Near Infrared (NIR) cameras are also used for driver gaze data collection \citep{nuevo2010rsmat, naqvi2018deep,  yoon2019driver} due to its intrinsic advantage to capture better features in night compared to traditional RGB cameras. It provides a gray-scale image using infrared/near-infrared light. Although NIR cameras are robust enough to capture images in low-light conditions, however, prolonged use of the NIR camera may hurt the driver's eyes \citep{ou2021deep}. The basic challenge in driver gaze estimation is the illumination vulnerability under poor environmental conditions where light and shade bring negative effects. Standard RGB cameras have the advantage of color information but are missing depth information. To overcome these challenges, RGB-D cameras \citep{ribeiro2019driver, wang2019continuous, ortega2020dmd, liu2022review} have been used to obtain RGB images and depth information using point cloud-based sensors. RGB-D camera has unique features to merge pixel-to-pixel information of depth and RGB information in a single image. Depth information of the camera is provided by a 3D depth sensor, which can be stereo, time of flight, structured light sensor, etc. 
In some studies, eye trackers are also used to capture data such as pupil dilation, iris, and gaze information, in terms of fixation and saccades \citep{palazzi2018predicting}.

\subsection{Collection methodology}
\begin{figure}
    \centering
    \includegraphics[width= 0.9\textwidth]{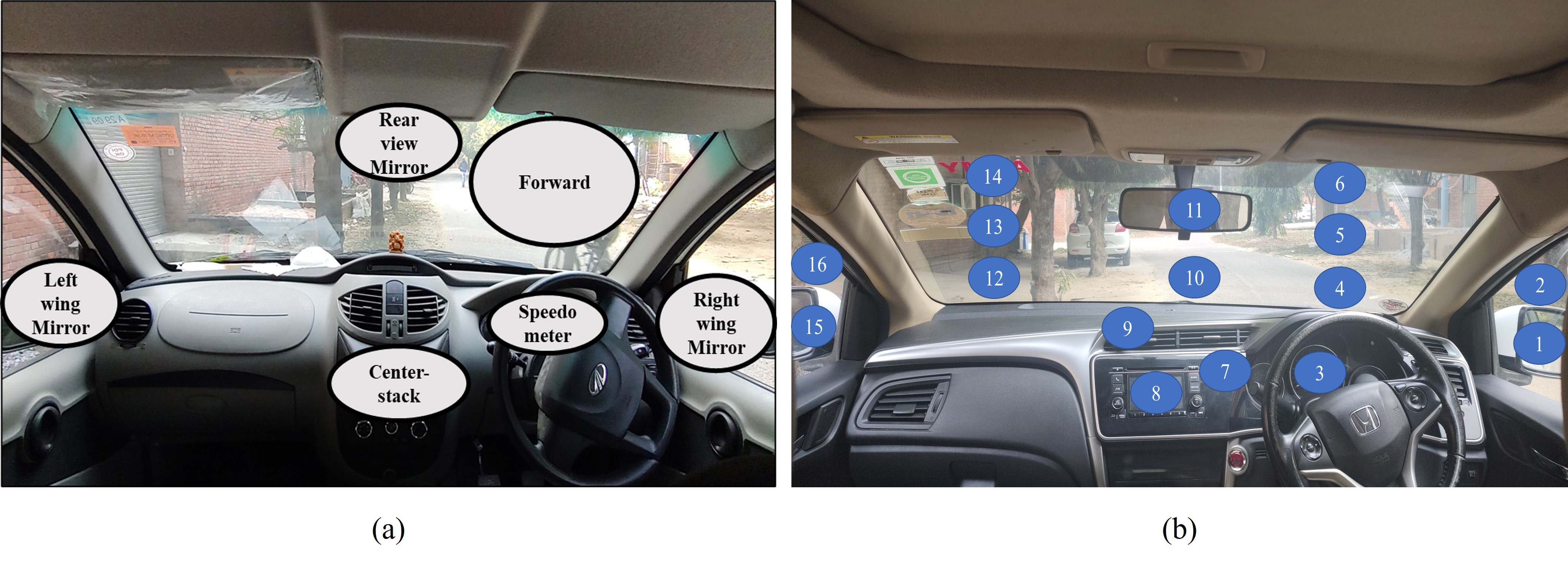}
    \caption{Gaze zone classification. (a) Coarser gaze zones (b) Finer gaze zones.}
    \label{fig:Gaze zone classification}
\end{figure}

Driver gaze data can be collected using a vehicle in stationary state (parked vehicle) or moving state. In the stationary state \citep{chuang2014estimating, naqvi2018deep, jha2018probabilistic, yoon2019driver, ribeiro2019driver, ghosh2021speak2label}, dashboard and windshield areas of the vehicle (car) are divided into several zones by sticking stickers or by pointing with a marker. The number of gaze zones can be broadly divided into coarser and finer categories, as shown in Fig. \ref{fig:Gaze zone classification}a and Fig. \ref{fig:Gaze zone classification}b. In the coarser gaze zone classification, the windshield and dashboard area are divided into fewer gaze zones than the finer. Primarily coarser gaze zone classification includes forward, speedometer, center-stack, left wing mirror, right wing mirror, and rearview mirror \citep{nuevo2010rsmat, chuang2014estimating, tawari2014attention, fridman2016driver,   martin2018dynamics, vora2018driver, rangesh2020driver}. In finer gaze zone classification, the windshield, wing mirror, and dashboard areas are subdivided into smaller gaze zones, resulting in more gaze zones.
Despite giving the specific class such as forward, rearview mirror, left-wing mirror, right-wing mirror, etc., they divided each broader gaze zone area into a smaller gaze zone with the class a numerical value such as 1, 2, 3, etc. \citep{vicente2015driver, naqvi2018deep, yoon2019driver, ghosh2021speak2label, wang2022driver} as shown in Fig. \ref{fig:Gaze zone classification}b. In some studies, instructions are given to the driver by a second person to the subjects to look toward specific gaze zones \citep{ribeiro2019driver}, while in some other studies, drivers typically look toward the gaze zones by their own choice \citep{chuang2014estimating, ortega2020dmd}. The frontal face area of the driver is recorded by installing the camera on the dashboard, rearview mirror, or windshield (see Fig. \ref{fig: Gaze zone estimation techniques}b). One or more human annotators label the captured frames and cross-verify their labels with each other. Speak2label \citep{ghosh2021speak2label} is another method of annotations in which audio signals (gaze zone class such as 1, 2, 3, etc.) are converted into text. 
In the moving state, we can not instruct the driver to look toward the specific gaze zone due to the risk of driver safety concerns. In this state, the driver drives the vehicle actually on the road, and the labels are given either by the human annotators \citep{fridman2016owl, vora2018driver, wang2019continuous}, using eye-trackers, or using unsupervised machine learning (ML) techniques \citep{chuang2014estimating}. 
\par The primary advantage of stationary vehicle-based data collection methods is the associated safety and more control during data collection since data is collected on parked vehicles. Getting the gaze zone label using speak2label or looking towards the specific gaze zone based on the instructions given by another person is relatively smooth. The drawback of this method is the gaze zone classifiers built using this data are not generalized satisfactorily from a stationary to a moving vehicle. Also, some classes can be intermingled thereby affecting the classifier results. The moving vehicle based data collection method is more generalized since data is collected in actual road driving, however, interpreting the correct gaze zone labels is challenging.

\subsection{Driver gaze datasets}
Driver gaze datasets can divided into two categories based on the details of the information captured. The first category contains driver face information, and the second category includes the eye pose information, such as the position of the iris, pupil, and cornea reflection inside the eye. Typically, driver face and eye data are collected using cameras \citep{Fusek2018433} while iris and pupil data are collected using eye trackers \citep{palazzi2018predicting}.
\subsubsection{Driver face benchmark datasets}
Several open-source driver face datasets are available, which are collected either inside parked vehicles or moving vehicles in real-world conditions. These datasets can be downloaded from different open-source repositories or available on request through a source generator. A detailed description of these publicly available driver gaze datasets is discussed next, and a comprehensive summary is provided in Table \ref{table:1}. 
\par
\begin{figure}
    \centering
    \includegraphics[width= 0.9\textwidth]{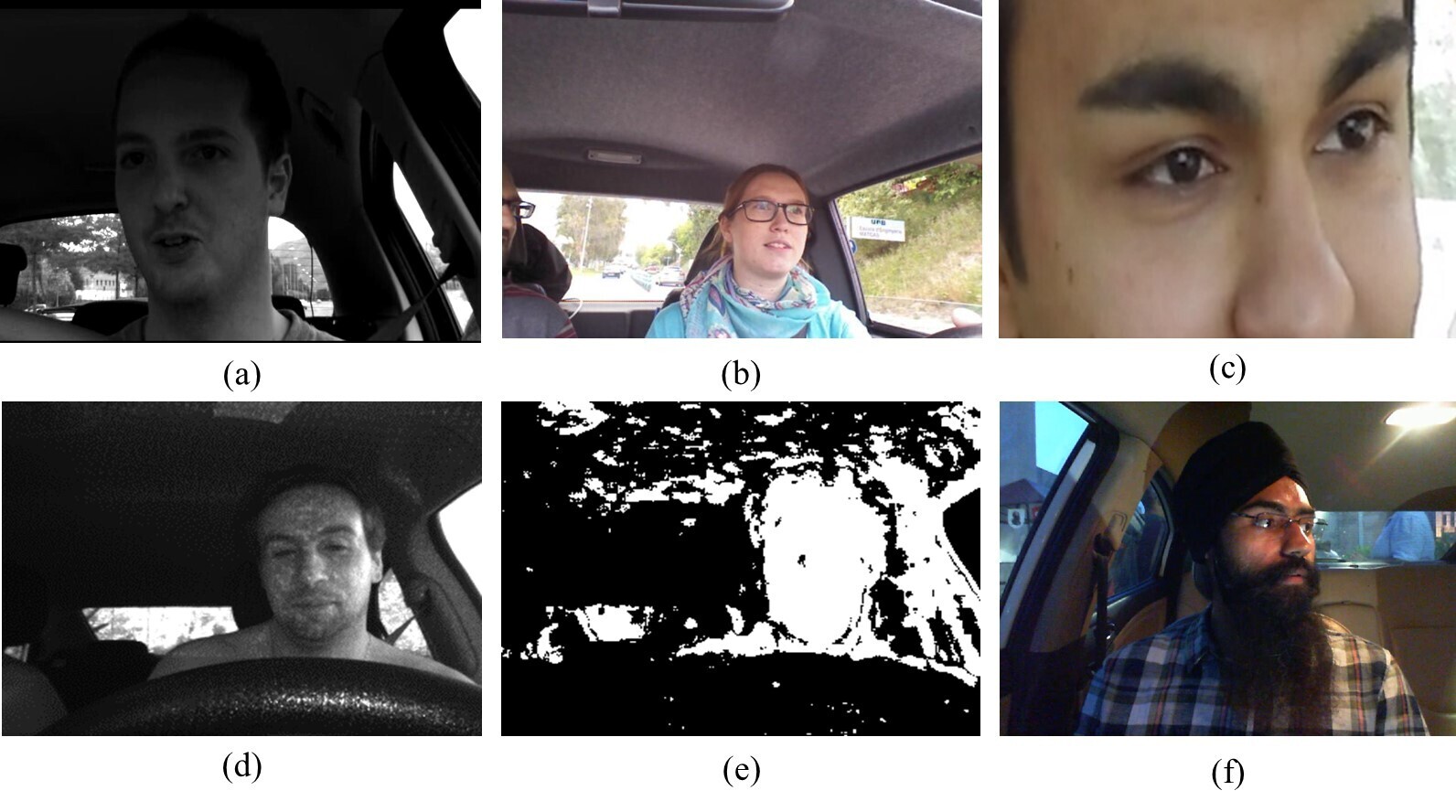}
    \caption{Open source benchmark driver's face datasets image samples. (a) RS-DMV \citep{nuevo2010rsmat} (b) DriveFace \citep{diaz2016reduced} (c) LISA GAZE v2 \citep{rangesh2020driver} (d)DG-UNICAMP \citep{ribeiro2019driver} (e) DG-UNICAMP \citep{ribeiro2019driver} (f) DGW \citep{dua2020dgaze}. }
    \label{fig:Open source Datset}
\end{figure}

 RS-DMV \citep{nuevo2010rsmat} contains grayscale face videos of ten drivers (see Fig. \ref{fig:Open source Datset}a) in indoor (simulator) and on-campus outdoor driving. DriveFace \citep{diaz2016reduced} contains three classes: right, frontal, and left head pose (see Fig. \ref{fig:Open source Datset}b). Brain4Cars \citep{Jain_2015_ICCV} consists of multi-sensor synchronized data containing outside and inside views of car, vehicle speed, and GPS coordinates. The data was recorded using ten drivers in natural driving settings for up to two months. DriveAHEAD \citep{schwarz2017driveahead} dataset is a wide-range head pose dataset containing depth and IR images. To measure the head position (x, y, z coordinate) and orientation (yaw, pitch, and roll), they used a 3D motion capture sensor. The DMD \citep{ortega2020dmd} dataset is a multimodal dataset containing images from 3 camera streams (RGB, IR, Depth). This dataset contains head pose, body pose, blink rate, and hand wheel interactions. DriveMVT \citep{othman2022drivermvt} is a multi-purpose natural driving data consisting of frame-by-frame information of driver health, such as heart rate, mental fatigue, head pose (yaw, pitch, roll), drowsiness, etc. The data was collected using USB cameras and smartphone cameras to capture driver face, while heart rate recording was obtained uing Xiaomi Mi Band 3 sensor. All these datasets were collected when the vehicle was in movement state.

\par Driver gaze datasets are also recorded inside parked vehicles. One such dataset is LISA GAZE v2 \citep{vora2018driver}, a large-scale driver face data containing diverse driving conditions such as daylight, night light, harsh illumination, and eyeglasses reflections (see Fig. \ref{fig:Open source Datset}c). DG-UNICAMP \citep{ribeiro2019driver} is another extensive driver face dataset containing three camera types RGB, IR (see Fig. \ref{fig:Open source Datset}d), and depth (see Fig. \ref{fig:Open source Datset}e) images. Driver Gaze in Wild (DGW) dataset \citep{ghosh2021speak2label} is one of the largest datasets in terms of the participants involved. This study used speak-to-label technique to label data. It contains different challenging lighting conditions, such as low light (see Fig. \ref{fig:Open source Datset}f), half/full face shadow, sunlight reflection, etc. The drawback of the data is its intermingled gaze classes, which can reduce the gaze classification accuracy.  

\begin{table}[!htb]
\fontsize{8pt}{8pt}\selectfont
\caption{Open source driver gaze datasets.}
\captionsetup[Table]{position=top}
\begin{tabular}{llllllll}
\hline
\begin{tabular}[c]{@{}l@{}}Name \end{tabular}         & \begin{tabular}[c]{@{}l@{}}Camera \\ type\end{tabular}     & Resolution & \begin{tabular}[c]{@{}l@{}}No. of \\ subjects\\ (F/M)\end{tabular} & \begin{tabular}[c]{@{}l@{}}Gaze \\ zones\end{tabular} & Size & Merits                                                                                                                               & Demerits                                                                                                                       \\ \hline
\begin{tabular}[c]{@{}l@{}}RS-DMV \\ \citep{nuevo2010rsmat} \end{tabular}       & Grayscale                                                  & 960×240    & \begin{tabular}[c]{@{}l@{}}10 \\ (2/8)\end{tabular}                & -                                                     & 21k  & \begin{tabular}[c]{@{}l@{}}Both indoor \\ and outdoor \\ environments\end{tabular}                                                   & \begin{tabular}[c]{@{}l@{}}Only grayscale \\ images\end{tabular}                                                               \\
\begin{tabular}[c]{@{}l@{}}DrivFace\\  \citep{diaz2016reduced}\end{tabular}     & RGB                                                        & 640×480    & \begin{tabular}[c]{@{}l@{}}4 \\ (2/2)\end{tabular}                 & 3                                                     & 606   & \begin{tabular}[c]{@{}l@{}}Real driving \\ scenarios\end{tabular}                                                                    & \begin{tabular}[c]{@{}l@{}}Limited number of\\ subjects, invariability\\ in lighting condition\end{tabular}                    \\
\begin{tabular}[c]{@{}l@{}}Brain4Cars\\ \citep{Jain_2015_ICCV}\end{tabular}   & RGB                                                        & -          & \begin{tabular}[c]{@{}l@{}}10 \\ (3/7)\end{tabular}                & -                                                     & 2M     & \begin{tabular}[c]{@{}l@{}}1180 miles \\ city and freeway \\ real driving data, \\ with with GPS, \\ IMU, camera \\ \end{tabular}        & \begin{tabular}[c]{@{}l@{}}Limited number \\ of subjects\end{tabular}                                                          \\
\addlinespace
\begin{tabular}[c]{@{}l@{}}DriveAHEAD\\ \citep{schwarz2017driveahead}\end{tabular}   & \begin{tabular}[c]{@{}l@{}}Depth\\ /IR\end{tabular}        & 512×424    & \begin{tabular}[c]{@{}l@{}}24 \\ (4/16)\end{tabular}               & -                                                     & 1M    & \begin{tabular}[c]{@{}l@{}}Real driving \\ scenarios\end{tabular}                                                        &  \begin{tabular}[c]{@{}l@{}}Only depth, \\ IR images \end{tabular}                 \\
\addlinespace
\begin{tabular}[c]{@{}l@{}}DR(eye)VE\\ \citep{palazzi2018predicting}\end{tabular}    & RGB                                                        & 1280×720   & \begin{tabular}[c]{@{}l@{}}8 \\ (1/7)\end{tabular}                 & -                                                     & 555k  & \begin{tabular}[c]{@{}l@{}}Real driving \\ scenarios with \\ different lighting \\ conditions\end{tabular}                           & \begin{tabular}[c]{@{}l@{}}Limited number \\ of subjects\end{tabular}                                                          \\
\addlinespace
\begin{tabular}[c]{@{}l@{}}LISA GAZE v2\\ \citep{rangesh2020driver}\end{tabular} & RGB                                                        & 2704×1524  & \begin{tabular}[c]{@{}l@{}}10 \\ (4/6)\end{tabular}                & 7                                                     & 47k   & \begin{tabular}[c]{@{}l@{}}Different\\ lighting \\ conditions\end{tabular}                                                           & Simulator driving                                                                                                              \\
\addlinespace
\begin{tabular}[c]{@{}l@{}}DG-UNICAMP\\ \citep{ribeiro2019driver}\end{tabular}   & \begin{tabular}[c]{@{}l@{}}RGB\\ /IR\\ /Depth\end{tabular} & 240×320    & \begin{tabular}[c]{@{}l@{}}45 \\ (10/35)\end{tabular}              & 18                                                    & 1M    & \begin{tabular}[c]{@{}l@{}}First driver face \\ dataset which \\ contains all three \\ cameras type \\ (RGB, IR, depth)\end{tabular} & Stationary vehicle                                                                                                             \\
\addlinespace
\begin{tabular}[c]{@{}l@{}}DGW\\ \citep{ghosh2021speak2label}\end{tabular}          & RGB                                                        & 179×179    & \begin{tabular}[c]{@{}l@{}}338 \\ (91/247)\end{tabular}            & 9                                                     & 50k   & \begin{tabular}[c]{@{}l@{}}Large number of        \\ subjects with \\ different lighting \\ conditions\end{tabular}                  & Stationary vehicle                                                                                                             \\
\addlinespace
\begin{tabular}[c]{@{}l@{}}DMD\\ \citep{ortega2020dmd}\end{tabular}          & \begin{tabular}[c]{@{}l@{}}RGB\\ /IR\\ /Depth\end{tabular} & 1920×1080  & \begin{tabular}[c]{@{}l@{}}37 \\ (10/27)\end{tabular}              & 9                                                     & 41h   & \begin{tabular}[c]{@{}l@{}}Both real world and \\ simulated driving \\ scenarios\end{tabular}                                        & \begin{tabular}[c]{@{}l@{}}Subjects performed \\ predefined activities \\ such as drinking water, \\ texting etc.\end{tabular} \\
\addlinespace
\begin{tabular}[c]{@{}l@{}}DGAZE\\ \citep{dua2020dgaze}\end{tabular}        & RGB                                                        & 1920×1080  & \begin{tabular}[c]{@{}l@{}}20\\ (6/14)\end{tabular}                & -                                                     & 100k  & \begin{tabular}[c]{@{}l@{}}Gaze dataset on the \\ vehicle entities \\ e.g. cars, motorcycles,\\ etc.\end{tabular}                    & Simulator driving                                                                                                              \\ \hline
\end{tabular}
\label{table:1}
\end{table}

\subsubsection{Open source driver eyes tracking datasets}
While the driver face dataset is the most common driver gaze dataset explored in literature, a few studies have also used the driver eye dataset for gaze estimation. Besides gaze estimation tasks, driver eye datasets are also used for detecting drowsiness, pupil dilation, and blink frequency for cognitive workload, etc. 
\par DR(eye)VE \citep{palazzi2018predicting}, is a real-world driving dataset consisting of 74 videos of different weather (sunny, cloudy, rainy) and light conditions (day, evening, night). Driver gaze information and pupil dilation were captured using eye trackers, and the gaze was mapped to the surrounding traffic. Media research lab (MRL) \citep{Fusek2018433} dataset was recorded on actual road driving using a NIR camera to reduce the low illumination light effect on the eyes during evening and night. IR illuminator is used to create reflection on eyes and eyeglasses to produce a wide range of lighting effects. The benchmark driver face and eye datasets are used to build several state-of-the-art gaze estimation models, which are discussed in the next section.

\section{Algorithms and Models for Driver Gaze Estimation}
Existing driver gaze estimation techniques can be divided into two groups: Appearance-based methods and Geometric model-based methods. Predominantly, appearance-based gaze estimation techniques have been used in driver gaze estimation; however, some studies have also used geometric model-based techniques. Appearance-based methods can be further subdivided into two groups: conventional appearance-based methods and appearance-based methods with deep learning (DL). 
\par This section will discuss the appearance and geometric model-based methods, along with their strengths and shortcomings. We will start with discussion of different features and traditional machine learning classifiers  (e.g., support vector machine etc.) used in conventional appearance-based methods. Then, we will discuss appearance-based deep learning methods, mainly focusing on convolutional neural network (CNN) models used in driver gaze estimation. Finally, we will highlight some studies that use geometrical models to estimate driver gaze. 

\subsection{Appearance based methods}
Appearance-based methods derive insights from the visual appearance of driver face and eyes. Conventional appearance-based methods rely on feature extractors to extract the features, while appearance-based deep learning methods learn the features directly through the convolutional layers.

\subsubsection{Conventional appearance based method}
Conventional appearance-based gaze estimation methods rely on handcrafted features and traditional machine learning models to estimate driver gaze. The features, such as texture, shape, edges, etc. are extracted from the eyes and face region.

\par In conventional appearance-based methods, head pose and eye pose are the two important aspects of driver gaze estimation. Different features are extracted from the driver facial image for head pose and eye pose. The common features for head pose are Euler angle (Yaw, Pitch, and Roll) \citep{lee2011real, tawari2014attention, tawari2014robust, wang2019continuous} or the facial landmark (such as mouth and nose corner, nose tip, eyebrows, etc.) \citep{lee2011real, chuang2014estimating, tawari2014attention, tawari2014robust}. On the other hand, the commonly used features extracted from the eye region are eye corners, eye contours, pupil, iris center location,  \citep{chuang2014estimating, tawari2014attention, fridman2016owl}. 
The extracted features using different feature extractors are then fed to conventional machine learning models for regression based gaze direction estimation or gaze zone classification. The commonly used traditional machine learning models are Support Vector Machine (SVM) \citep{lee2011real, chuang2014estimating, vicente2015driver, vasli2016driver}, Random Forest (RF) \citep{tawari2014robust, tawari2014attention, fridman2016owl, fridman2016driver, wang2019continuous}, etc.  We will first discuss the driver gaze studies where the gaze output is gaze zone class such as forward, rearview mirror, etc. (see Fig. \ref{fig:Gaze zone classification}), followed by the studies where the gaze output is point of gaze.

For driver gaze estimation based on gaze zone class, \citet{chuang2014estimating} estimated coarse head pose direction (frontal, left, and right) using Adaboost cascade classifiers operating on Haar feature descriptors. Then, only the frontal face region was fed to extract the location of the left iris, right iris, mouth, and nose using  Adaboost cascade classifier. Finally, the 14 feature vectors comprising of face part location were fed to the ML model (as demonstrated in Table \ref{table:2})  for the gaze estimation. 
Study by \citet{tawari2014robust} incorporated both static head pose features and dynamic head pose features for gaze estimation. Static head pose features based on facial landmarks and their 3D configuration were determined through a perspective projection model. The corresponding yaw, pitch, and roll angles were then obtained using POS (pose from orthography and scaling).
For dynamic head pose feature extraction, raw data from static features were first smoothed. From the windowed time series (spanning W seconds prior to the current time \textit{t}), min and max values, positions of min and max, mean angle, and angular velocity were extracted. The total 18 dynamic features were then fed into the classifier to estimate the gaze zone class. However, since this study used only the head pose, the trained model confused between some nearby classes, such as forward and the speedometer, center console, and rearview mirror. Therefore, in another study, \citet{tawari2014driver}
added eye pose features (iris position) to improve accuracy.  \citet{fridman2016owl} tested how much the gaze estimation accuracy increased when the eye pose was included along with the head pose. For the head pose, 68 points of multi-PIE facial landmark markup were extracted, while for the eye pose, the pupil center was extracted by extracting points of iso-contour. Adding eye pose information with head pose was found to increase 5.4 \% accuracy of the gaze zone classification system.

In the literature, when driver gaze output is a gaze zone class, RF is preferred over SVM and KNN for several reasons. RF consistently exhibits higher classification accuracy, as demonstrated in Table \ref{table:2}. Additionally, RF provides prediction probabilities for each class and requires very few tuning parameters.

\par
\begin{figure}[!htb]
    \centering
   \includegraphics[width=0.95\textwidth]{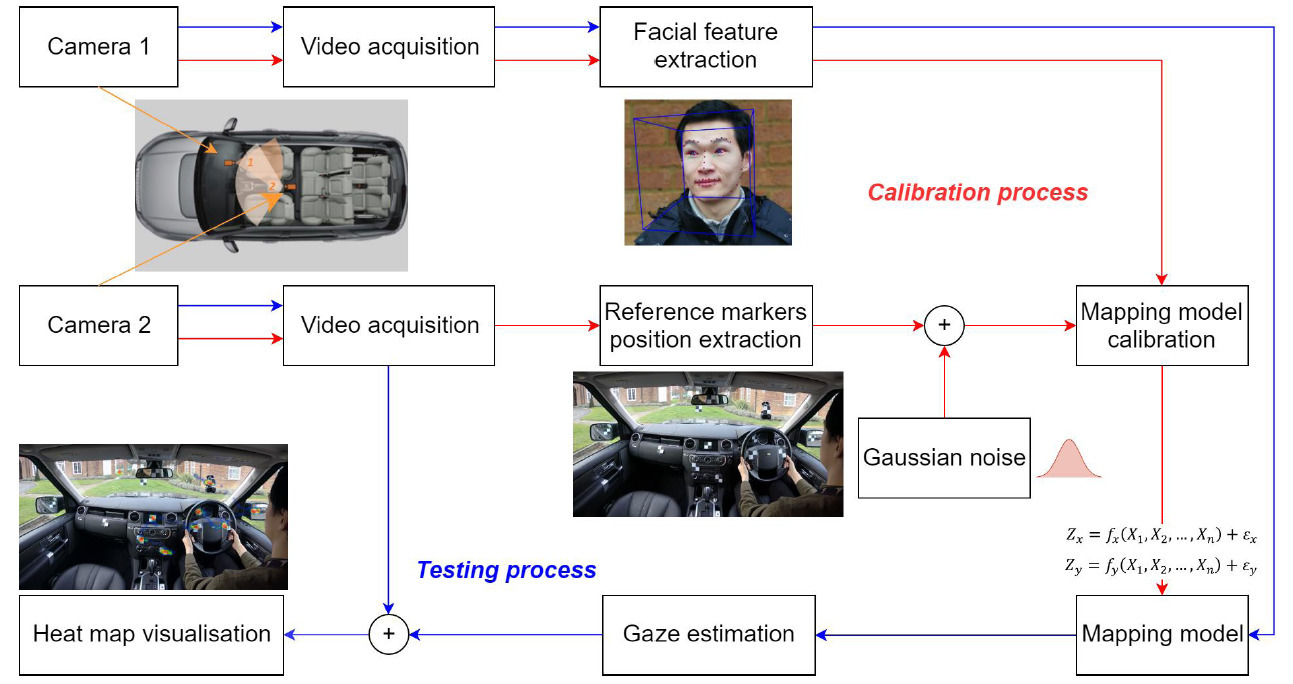}
    \caption{Systematic flow of driver gaze mapping \citep{yang2019dual} using dual camera. Camera 1: Capture the facial features including eye gaze and head movement. Camera 2: Used to mapped location of eye gaze in scene images.}
    \label{fig:Gaze Mapping}
\end{figure}

 While a significant number of studies have been conducted based on gaze zone class, limited research have been done on gaze point location. Study by \citet{yang2019dual} used dual cameras for driver gaze mapping, as shown in Fig. \ref{fig:Gaze Mapping}.  Camera 1 was placed on the front of the driver on the dashboard to capture the facial features, and camera 2 was placed at the driver's top to capture the driver scene images.  A nonlinear finite impulse response (NFIR) model was utilized to depict a system with several inputs and a single output where the inputs were the face features and the output was the point of gaze on camera 2 image. {\color{black}To increase the accuracy of driver gaze estimation,  \citet{yuan2022self} presented a self-calibrated driver gaze estimation system. Features extraction, gaze pattern learning, and gaze mapping model training comprised of three stages. Using OpenFace, they initially extracted features like head and eye poses and facial landmarks, which were tracked using unscented Kalman filter. In the gaze pattern learning step, unaligned typical gaze patterns with gaze features were automatically extracted from the time samples. In the third step, the gaze was mapped using Gaussian Processes for Regression (GPR), which builds the relationship between driver status and point of gaze.}

\subsubsection{Appearance based method with deep learning}
Recently, deep learning-based driver gaze estimation has gained significant attention. When contrasted with conventional appearance-based methods, these methods exhibit several notable advantages. Deep learning based methods directly learn the mapping functions from the face and eye appearance characterized by pixel intensity (color intensity) statistics \citep{lrd2022distraction, pathirana2022eye}. Typically, they do not require handcrafted feature extractors. These methods work well in real-world settings, both in head-mounted and remote setup-based gaze estimation. They are robust for head movements and also improve cross-subject gaze estimation performance. Next, we will discuss different deep learning-based driver gaze estimation models. Initially, we focus on CNN models used in driver gaze estimation based on gaze zone classification. Later, we will discuss the studies where estimated gaze output is point of gaze.

\begin{table}[!htb]
\begin{threeparttable}
\fontsize{8pt}{8pt}\selectfont
\caption{Gaze estimation using conventional appearance based method.}
\captionsetup[Table]{position=top}
\begin{tabular}{lllllll}
\hline
Reference                                                       & \begin{tabular}[c]{@{}l@{}}No. of \\ cameras \\ (type)\end{tabular} & \begin{tabular}[c]{@{}l@{}}Output\\ (no. of \\ zones)\end{tabular} & Features                                                                                                                                            & \begin{tabular}[c]{@{}l@{}}Feature \\ extractor\end{tabular}                                                                                       & \multicolumn{1}{c}{Model} & \multicolumn{1}{c}{\begin{tabular}[c]{@{}c@{}}Accuracy (\% or °)\end{tabular}} \\ \hline
\begin{tabular}[c]{@{}l@{}}  \citet{lee2011real}\end{tabular}                                               & 1 (NIR)                                                             & \begin{tabular}[c]{@{}l@{}}GZC\tnote{1}\\(18) \end{tabular}                                                           & \begin{tabular}[c]{@{}l@{}}Head pose (yaw, pitch), \\ face features \\ (Left, right border and \\ center of drivers face)\end{tabular}              & \begin{tabular}[c]{@{}l@{}}Head pose by \\ support vector \\ regressor (SVR)\end{tabular}                                                          & SVM                       & \begin{tabular}[c]{@{}l@{}}47.4 (SCER),  \\ 87.2 (LCER)\end{tabular}     \\
\begin{tabular}[c]{@{}l@{}}  \citet{chuang2014estimating}\end{tabular}                                                 & 1 (RGB)                                                             & \begin{tabular}[c]{@{}l@{}}GZC\\(2) \end{tabular}                                                                & \begin{tabular}[c]{@{}l@{}}Head pose and eye pose\\ (location of left, right \\ iris, mouth and nose)\end{tabular}                                  & \begin{tabular}[c]{@{}l@{}}Haar featur \\ descriptor, \\ haar cascade \\ classifier\end{tabular}                                                   & SVM                       &\begin{tabular}[c]{@{}l@{}}97.4\end{tabular}                                                                             \\
\begin{tabular}[c]{@{}l@{}}  \citet{tawari2014robust}\end{tabular} & \begin{tabular}[c]{@{}l@{}} 2 (RGB)\end{tabular} & \begin{tabular}[c]{@{}l@{}} GZC\\(8)\end{tabular} &\begin{tabular}[c]{@{}l@{}}Facial land marks \\ (eye and nose corner, \\ nose tip), static features\\ of head pose (yaw,\\ pitch, roll)\end{tabular} & \begin{tabular}[c]{@{}l@{}}\\ WPPM\tnote{2}\hspace{3pt}  for \\landmark tracking,\\POS\tnote{3} \hspace{2pt} for determines \\the rotation matrix\\and corresponding\\ yaw, pitch and roll\end{tabular} & RF  & \begin{tabular}[c]{@{}l@{}}94.7\end{tabular}  \\

\begin{tabular}[c]{@{}l@{}} \citet{tawari2014attention}\end{tabular}                                                 & 2 (RGB)                                                             & \begin{tabular}[c]{@{}l@{}}GZC\\(6) \end{tabular}                                                                 & \begin{tabular}[c]{@{}l@{}}Head pose (yaw, pitch, \\roll),  eye pose (by pupil), \\features such as eye \\corners and contours\end{tabular}    & \begin{tabular}[c]{@{}l@{}}\\ WPPM for \\landmark tracking,\\POS for determines \\the rotation matrix\\and corresponding\\ yaw, pitch and roll\end{tabular}  & RF                        & \begin{tabular}[c]{@{}l@{}}79.8 (H), \\ 94.9 (H + E)\tnote{4}\end{tabular}              \\
\addlinespace
\begin{tabular}[c]{@{}l@{}} \citet{vicente2015driver}\end{tabular}                                                & 1 (NIR)                                                             & \begin{tabular}[c]{@{}l@{}}GZC\\(18) \end{tabular}                                                                & \begin{tabular}[c]{@{}l@{}}Facial features (pupil \\and eye corners)\end{tabular}                                                                  & Shift descriptor                                                                                                                                   & SVM                       & \begin{tabular}[c]{@{}l@{}}90\end{tabular}                                                                                \\
\begin{tabular}[c]{@{}l@{}} \citet{vasli2016driver}\end{tabular}                                                  & 1 (RGB)                                                             & \begin{tabular}[c]{@{}l@{}}GZC\\(6) \end{tabular}                                                                & \begin{tabular}[c]{@{}l@{}}Head pose\\ (yaw, pitch, roll)\end{tabular}                                                                           & \begin{tabular}[c]{@{}l@{}}Face detection \\ using AlexNet \\ model\end{tabular}                                                                   & SVM                       & \begin{tabular}[c]{@{}l@{}}93.7\end{tabular}                                                                              \\
\addlinespace
\begin{tabular}[c]{@{}l@{}} \citet{fridman2016owl}\end{tabular}                                                 & 1 (NIR)                                                             & \begin{tabular}[c]{@{}l@{}}GZC\\(6) \end{tabular}                                                                & \begin{tabular}[c]{@{}l@{}}3D Head pose, eye pose \\ by pupil detection.\end{tabular}                                                               & \begin{tabular}[c]{@{}l@{}}DLIB HOG face \\ detection, pupil \\ center detection \\ based on CDF\tnote{5}\end{tabular}                                      & RF                        & \begin{tabular}[c]{@{}l@{}} 89.2 (H)\\94.6 (H+E)\end{tabular}                \\
\addlinespace
\begin{tabular}[c]{@{}l@{}}  \citet{fridman2016driver}\end{tabular}                                                 & 1 (NIR)                                                             & \begin{tabular}[c]{@{}l@{}}GZC\\(6) \end{tabular}                                                              & \begin{tabular}[c]{@{}l@{}}19 facial landmarks \\ e.g. nose tip, eye \\ corner etc.\end{tabular}                                                    & \begin{tabular}[c]{@{}l@{}}DLIB HOG face \\ detection, face \\ alignment using\\  68 Multi-PIE \\ facial landmark\end{tabular}                     & RF                        & \begin{tabular}[c]{@{}l@{}}91.4\end{tabular}                                                                              \\
\addlinespace
\begin{tabular}[c]{@{}l@{}} \citet{wang2019continuous}\end{tabular}                                                    & 1 (RGB-D)                                                           & \begin{tabular}[c]{@{}l@{}}PoG\tnote{6}\end{tabular}                                                              & \begin{tabular}[c]{@{}l@{}}Head pose (yaw, pitch,\\ roll), localized eye \\region, eye pose\end{tabular}                                        & \begin{tabular}[c]{@{}l@{}}Multi-zone ICP\tnote{7}\hspace{2pt} \\ based head pose  \\ and appearance \\ based gaze\end{tabular}                                         & RF                        & \begin{tabular}[c]{@{}l@{}}8.0°\end{tabular}                                                                          \\
\addlinespace
\begin{tabular}[c]{@{}l@{}} \citet{yang2019dual}\end{tabular}                                                    & 2 (RGB)                                                           &  \begin{tabular}[c]{@{}l@{}}PoG\end{tabular}                                                      & \begin{tabular}[c]{@{}l@{}} 68 facial landmarks \\  head pose, eye pose\end{tabular}                                        & \begin{tabular}[c]{@{}l@{}}OpenFace for face \\detection, conditional \\ local neural fields \\(CLNF) for facial \\landmark detection\end{tabular}                                         & NFIR\tnote{8}                        &\begin{tabular}[c]{@{}l@{}} RMS\tnote{9} \end{tabular} \\ 
\addlinespace
\begin{tabular}[c]{@{}l@{}} \citet{yuan2022self}\end{tabular}                                                    & 1 (RGB)                                                           &  \begin{tabular}[c]{@{}l@{}}PoG\\(8)\end{tabular}                                                      & \begin{tabular}[c]{@{}l@{}} 68 facial landmarks \\  head pose, eye pose\end{tabular}                                        & \begin{tabular}[c]{@{}l@{}}OpenFace,  \\Unscented Kalman filter \\(for tracking the feature \\ over time)\end{tabular}                                         & GPR                       &\begin{tabular}[c]{@{}l@{}}6.9° \end{tabular} \\ \hline
\end{tabular}
\begin{tablenotes}[para,flushleft]
\item[1] Gaze zone class;\item[2] Weak perspective projection model;\item[3] Pose from orthography and scaling;\item[4] H-Head pose, E-Eye pose;\item[5]Cumulative distribution function;\item[6] Point of gaze;\item[7] Iterative closet points;\item[8] Non linear finite impulse response;\item[9] Root mean square error used to measurement of gaze estimation error, in X and Y direction in terms of  pixels are ((7.80$\pm$5.99)) \hspace{2pt}\textit{{\normalfont and}}\hspace{2pt} (4.64$\pm$3.47).
\end{tablenotes}
\label{table:2}
\end{threeparttable}
\end{table}

\paragraph{CNN models for driver gaze estimation}
Several state-of-the-art CNN-based models mentioned in Table \ref{table:3} have been used for driver gaze estimation. These models are generally pre-trained on the large-scale ImageNet dataset \citep{deng2009imagenet} and applied to driver face data, collected using RGB or NIR cameras. We first discuss studies that used RGB camera data, followed by NIR camera based studies, where the estimated gaze output is the gaze zone class. \citet{choi2016real} proposed a CNN model inspired by Alexnet architecture \citep{NIPS2012_c399862d}, consisting of three convolutional, three pooling, two fully connected, and one output layer for nine gaze zone classifications. Similarly, several other studies have focused on leveraging the advantage of large-scale CNN models by developing gaze classification models that can generalize for different drivers, driver position and perspective, lighting conditions, etc., as shown in Table \ref{table:3}. \citet{vora2018driver} collected gaze data of ten drivers in naturalistic driving conditions during dry weather in the daytime to build a generalized driver gaze classification model using CNN architecture. They achieved the highest accuracy of 88.9 \% and 93.4 \% accuracy in half-face data type by using AlexNet and VGG16, respectively. The most likely reason for getting higher accuracy on the upper half face compared to the full face data was that the upper half face images can extract finer features of the eyes, like the position and shape of the iris and eyelid. One of the challenges in gaze classification is reflection or low visibility due to eyeglasses. \citet{rangesh2020driver} attempted to overcome the eyeglass challenges by removing the eyeglasses in the real driving environment using the Gaze Preserving CycleGAN (GPCycleGAN) model. 

\par Gaze estimation using RGB cameras discussed above suffers during low light and sunlight reflection on the driver face. To solve this problem, researchers collected driver face data using NIR camera. \citet{naqvi2018deep} used three separate VGG models, one each for the driver's face, left eye, and right eye, respectively to estimate driver gaze. The accuracy of the proposed system was measured using two metrics: strictly correct estimation rate (SCER) and loosely correct estimation rate (LCER). SCER refers to the ratio of the number of strictly correct frames divided by the total number of frames. The strictly correct frame represents the frame where the estimated gaze zone is equivalent to the ground truth gaze zone. LCER refers to the ratio of the number of loosely correct frames divided by the total number of frames. The loosely correct frame represents the frame where the estimated gaze zone is placed within the ground truth gaze zone or its surrounding zones. The system achieved 92.8 \% and 99.6 \% accuracy in SCER and LCER, respectively. 
Using two NIR cameras, \citet{yoon2019driver} captured the drive frontal face and right side face images. Each face image was further used to detect left and right eyes using Dlib facial feature trackers. Finally, they combined six images and made a single image, giving it as a input to the CNN model for gaze estimation. 
 
\par All the above-discussed studies in this subsection estimated gaze zone class as the gaze output. However, few studies have also estimated driver gaze in terms of point of gaze. A study by \citet{lrd2022distraction} developed a system to estimate driver gaze using a modified Attention-based Gaze Estimation network. The estimated gaze error across eight gaze regions was 15.61 ± 10.4 cm in the x-direction and 15.13 ± 14.5 cm in the y-direction. 

\par Overall, deep learning-based CNN models have been preferred for driver gaze estimation because of their higher ability to incorporate large head movement and eye movement detection, robustness to variations in vehicle type and driver head \citep{kanade2021convolutional}, different lighting conditions such as low light, sunlight reflection on the face, and improved classification accuracy. These models have been demonstrated to be efficient in handling large-scale data, which helps in improving the model training to learn complex feature distribution and thereby finer gaze zone classification.

\begin{table}[!htb]
\begin{threeparttable}
\fontsize{8pt}{8pt}\selectfont   
\caption{Gaze estimation using appearance based method with deep learning}
\captionsetup[Table]{position=top}
\begin{tabular}{@{}llllll@{}}
\toprule
Reference            & \begin{tabular}[c]{@{}l@{}}No. of \\cameras \\ (type)\end{tabular} & \begin{tabular}[c]{@{}l@{}}Output \\(no. of \\ zones)\end{tabular} & Description                                                                                                                                                                                             & \begin{tabular}[c]{@{}l@{}}Best CNN\\ model\end{tabular}                           & \begin{tabular}[c]{@{}l@{}}Accuracy (\%)\end{tabular}                                                  
          \\ \midrule
\begin{tabular}[c]{@{}l@{}} \citet{choi2016real}\end{tabular}    & 1 (RGB)                                               & \begin{tabular}[c]{@{}l@{}}GZC\\(9)\end{tabular}                                                               & \begin{tabular}[c]{@{}l@{}}Face tracking using \\ MOSSE tracker, \\ compared with\\  Haar cascade classifier\end{tabular}                                                                               & AlexNet                                                                               & \begin{tabular}[c]{@{}l@{}}95.0\end{tabular}                                                                                \\
\addlinespace
\begin{tabular}[c]{@{}l@{}}  \citet{vora2018driver}\end{tabular}     & 1 (RGB)                                         &   \begin{tabular}[c]{@{}l@{}}GZC\\(7)\end{tabular}                                                              & \begin{tabular}[c]{@{}l@{}}Three separate data \\ with face, half face, \\ and face plus context\end{tabular}                                                                                            & \begin{tabular}[c]{@{}l@{}}SqueezeNet\end{tabular} & \begin{tabular}[c]{@{}l@{}}92.7\end{tabular}               \\
\addlinespace
\begin{tabular}[c]{@{}l@{}}  \citet{naqvi2018deep}\end{tabular}   & 1 (NIR)                                                   & \begin{tabular}[c]{@{}l@{}}GZC\\(17)\end{tabular}                                                              & \begin{tabular}[c]{@{}l@{}}Trained three separate \\ VGG16 on face,\\  left and right eye\end{tabular}                                                                                                  & VGG16                                                                                 & \begin{tabular}[c]{@{}l@{}}64.8 (SCER), \\ 91.1(LCER)\end{tabular}                \\
\addlinespace
\begin{tabular}[c]{@{}l@{}}  \citet{yoon2019driver}\end{tabular}      & 2 (NIR)                                                                & \begin{tabular}[c]{@{}l@{}}GZC\\(15)\end{tabular}                                                              & \begin{tabular}[c]{@{}l@{}}Generate single image \\ by combining 6 images of \\ frontal face, side face and \\ left and right eye of each face\end{tabular}                                             & \begin{tabular}[c]{@{}l@{}}ResNet50\end{tabular}        & \begin{tabular}[c]{@{}l@{}}92.9(SCER), \\ 99.5(LCER)\end{tabular} \\
\addlinespace
\begin{tabular}[c]{@{}l@{}} \citet{rangesh2020driver}\end{tabular} \ & 1 (IR)                                               & \begin{tabular}[c]{@{}l@{}}GZC\\(7)\end{tabular}                                                              & \begin{tabular}[c]{@{}l@{}}Include day and night time, \\ with and without eye glasses\\  images , remove eye glasses \\ using GPcycleGANE\end{tabular}                                                 & SqueezeNet                                                                            & \begin{tabular}[c]{@{}l@{}}72.4\end{tabular}                                                                            \\
\addlinespace
\begin{tabular}[c]{@{}l@{}} \citet{yang2021driver}\end{tabular}     & 1 (RGB)                                                  & \begin{tabular}[c]{@{}l@{}}GZC\\(7)\end{tabular}                                                                & \begin{tabular}[c]{@{}l@{}}Face detection using docker \\face, ERW-Net\tnote{10}\hspace{6pt}  for facial \\information from eyes\\ and mouth, HP-Net\tnote{11}\hspace{6pt} based on \\kronecker product mechanism \\for fusing unbalanced features \\of the face and head pose \end{tabular} & HP-ERW                                                           & \begin{tabular}[c]{@{}l@{}}72.11 (DGW) \\ 98.26 \\(LISA GAZE v1)\end{tabular}                                                                            \\
\addlinespace
\begin{tabular}[c]{@{}l@{}}  \citet{shah2022driver}\end{tabular}     & 1 (RGB)                                                 & \begin{tabular}[c]{@{}l@{}}GZC\\(7)\end{tabular}                                                              & \begin{tabular}[c]{@{}l@{}}YOLOV4 face detector \\ Inception-v3, robust feature \\ learning, improved \\ face detection\end{tabular} & InceptionResNet-v2                                                                              & \begin{tabular}[c]{@{}l@{}}91.0\end{tabular}                                                                                                                                  \\ \bottomrule           
\end{tabular}
\begin{tablenotes}[para,flushleft]
\item[10] Eye region weighted encoding network;\item[11] Head pose fusion network
\end{tablenotes}
\label{table:3}
\end{threeparttable}
\end{table}

\subsection{Geometric model based methods}

Geometric model based gaze estimation requires geometric model of an eye (see Fig. \ref{fig:3D eye model}) with eye features such as the center of the cornea, corneal reflection (glint),  pupil center (dark pupil or bright pupil), optical and visual axes of the eye \citep{akinyelu2020convolutional}. Geometric models use subject-specific parameters of geometric 3d eye models such as kappa angle. Because of variations between different drivers, kappa angle eye tracker requires calibration before using. Kappa angle is defined as the angle between the visual axis and the eye optical axis \citep{kar2017review, akinyelu2020convolutional}, which is used to estimate the gaze direction.
\par This method is typically used in studies based on head-mounted driver gaze estimation setup, where the driver eye information is captured from a closeup. However, some studies such as \cite{ji2001real, ji2002real, vicente2015driver, vasli2016driver} also used this method in remote setup. Studies by \cite{ji2001real, ji2002real} used eye pose and head pose for monitoring the driver by considering bright pupil effects. The pupil was tracked using Kalman filter with features around the pupil in conjunction with a KNN classifier for head pose estimation. Driver gaze was estimated using displacement between the center of the pupil to the glint with the help of a linear regression model to map nine gaze directions, thereby producing the final output as the nine gaze zone classes.
\par Geometric model-based methods offer notable advantages, including higher accuracy in gaze estimation and  ability to provide precise outputs in terms of point of gaze. These methods excel in controlled environments, which is why they are preferred in lab settings (simulation-based studies), delivering reliable results and maintaining low-latency responses. However, their application is tempered by certain limitations. Geometric models may exhibit limited robustness to individual variability, struggling to adapt seamlessly to differences in eye anatomy and facial characteristics among users. Challenges arise due to eye occlusion, where obscured visual cues hinder accurate gaze estimation. Sensitivity to calibration errors can also impact performance, emphasizing the importance of meticulous calibration procedures. In real-world scenarios, the complexity of environments, varying lighting conditions, and dynamic situations can pose challenges for geometric models, potentially compromising their accuracy.

\section{Uses and Application of Driver Gaze}
Driver gaze estimation is essential from several aspects. One critical aspect of driver gaze is understanding driver gaze behavior at different road sections, which helps to build safer road infrastructure and safety systems for drivers. Driver gaze is also used to build driver distraction detection system, inattention detection system, and advanced driver assistance system. This section is divided into two parts: first, we will discuss the uses of driver gaze in understanding driver behavior, followed by applications of driver gaze in real-world systems such as driver attentiveness detection, etc.

\subsection{Uses of driver gaze in driver behavior understanding}
Driver gaze behavior shows awareness of the driver to the surrounding traffic, such as vehicles coming from different traffic streams, road infrastructure such as traffic signs or traffic lights, road markings, etc., and roadside infrastructures such as buildings, trees, advertising hoardings, billboards, etc. Analyzing driver gaze behavior is crucial for evaluating attention to traffic signs and optimizing compliance with rules. Intersection design, stop sign, and traffic light placement impact gaze focus, with analysis aiding in enhancing design effectiveness. Gaze behavior assessment at intersections identifies potential conflict areas, helping in infrastructure improvements for increased safety. Road curvature, influenced by infrastructure design, affects driver navigation through curves. Gaze behavior analysis reveals drivers' adaptation to road geometries, offering insights for road design enhancements.
Roadside elements like billboards can divert driver attention. Monitoring gaze behavior helps understand driver interaction with these elements, assessing their impact on attention and safety. Driver gaze behavior analysis can be used by insurance companies to assess driver behavior and provide more accurate risk assessments, potentially leading to more personalized and fair insurance premiums. Therefore, it is critical to check if the drivers are aware of their surroundings before and while performing different traffic maneuvers such as lane changing, merging/diverging in on-ramp/off-ramps, and left/right turning at intersections. A detailed discussion of each aspect is provided in the following sections.

\subsubsection{Gaze behavior at intersections}
Intersections are known for their complex nature because of different participants' behaviors and interactions \citep{shirazi2016looking}. Interactions at intersection are vehicle-to-vehicle (V2V) \citep{harding2014vehicle, liu2018cooperation, ye2019deep}, vehicle-to-pedestrian (V2P) \citep{anaya2014vehicle, liu2018cooperation, sewalkar2019vehicle}, vehicle-to-infrastructure (V2I) \citep{milanes2012intelligent, liu2018cooperation}, and pedestrian-to-infrastructure (P2I) \citep{liu2018cooperation}. In the literature, interactions involving vehicles include drivers.  So the first three interactions are important from the driver's perspective and hence included in the present review. 

\par This section will discuss how driver gaze is influenced when the driver approaches, maneuvers (left turning, right turning, and going straight), and leaves the intersection. Literature on driver gaze behavior at intersections is broadly divided into three categories based on driver age or experience (novice, young experienced, old experienced), intersection types (signalized, unsignalized), and surrounding traffic environments (traffic density, familiarity), mentioned in Table \ref{table:4}.
 A few studies compared driver glance behavior based on their age or experience while approaching or negotiating through the intersection \citep{bao2009age, dukic2012older, scott2013visual, romoser2013comparing, savage2020effects, chung2022static}. The discussion will be based on left-hand driving to maintain uniformity in the paper.

\par In real driving scenarios, \citet{bao2009age} measured driver gaze scanning behavior and found that middle-aged drivers (35-55 years old) have higher scanning randomness (i.e., a greater entropy rate) than older drivers (65-80 years old). In another study, \citet{dukic2012older} found that older drivers looked more at road lines and markings to position themselves in surrounding traffic. Some other simulation-based studies examine the age and experience effect on driver gaze. While selecting a safe gap at an unsignalized intersection (USI), \citet{scott2013visual} compared glance transition patterns of three groups of drivers, including novice (mean age 20.57 years), young experienced (23.79 years),  and older experienced (66.43 years) on a right turn. They divided the intersection approach period into scanning phase (first 10 seconds before finding negotiable gaps) and decision phases (first 5 seconds after finding negotiable gaps), the results being shown in Table \ref{table:4}. A study by \citet{romoser2013comparing} examined four hypotheses to determine why older drivers fail to scan effectively at intersections compared to young drivers. The four hypotheses were difficulty with head movements, decrease in working memory capacity, increased distractibility, and failure to recall specific scanning patterns. None of the hypotheses were able to fully explain the above reason. Still, the research does support the alternative theory that some of the issues older drivers experience when looking at junctions are due to unique attentional weaknesses in the older drivers' ability. The effect of age and guidance type (lead car and GPS) \citet{savage2020effects} slightly reduced the gaze scanned when the driver is close to the intersection. \citet{chung2022static} compared the static (dwell time) and dynamic (gaze transitions) gaze of novice and experienced drivers. Static analysis of novice drivers shows higher dwell time in an area of interest (AOI) than experienced. 
 
\par The general observation from these studies was that older drivers, compared to younger drivers,  scan fewer right and left areas of interest and focus more straight ahead or in the intended direction of vehicles. This behavior may explain the fact that older drivers are more involved in angle crashes at intersections and "failure to yield" when involved in "seen but not seen" crashes and accidents with other vehicles \citep{stutts2009identifying, misra2023detection}.

\par In addition to studies focusing on driver scanning behavior in younger and older drivers, researchers have also focused on driver gaze behavior at SI and USI. \citet{li2019drivers} examined the influence of intersection types on driver scanning measures. When performing right-turning at SI, drivers were found to give more attention to the forward and right areas than at USI. 
 Two different studies, \citet{lemonnier2015gaze, lemonnier2020drivers} examined the impact of the three-factor priority rule (yield, priority, and stop), expected traffic density (no traffic, light, and heavy), and familiarity while approaching SI. The dwell time in intersecting road AOI was found to be higher in yield than in priority condition and smallest in stop sign condition. They also found that the horizontal gaze eccentricity is inversely related to traffic density. Horizontal eccentricity is defined as the absolute value of the horizontal component of the gaze (and head) direction angle. Visual information related to the decision-making task starts later when the driver is slightly familiar with the environment.

\begin{table}[!htbp]
\fontsize{8pt}{8pt}\selectfont
\caption{Driver gaze behavior at intersections.}
\captionsetup[Table]{position=top}
\begin{tabular}{lllll}
\hline
Reference                                                             & Study area                                                                                                               & \begin{tabular}[c]{@{}l@{}}Study type \\ and setup\end{tabular} & Gaze behavior                                                                                                                                                 & Key findings                                                                                                                                                                                                                                                                                 \\ \hline
\begin{tabular}[c]{@{}l@{}} \citet{bao2009age}\end{tabular}        & \begin{tabular}[c]{@{}l@{}}USI: left turn, \\ right turn, and \\ going straight\end{tabular}                               & \begin{tabular}[c]{@{}l@{}}Real driving \\(Remote \\ setup gaze \\ estimation)\end{tabular}     & \begin{tabular}[c]{@{}l@{}}Age related visual \\ difference in \\ proportion of time \\ glanced in each \\ gaze zone\end{tabular}                             & \begin{tabular}[c]{@{}l@{}}During intersection negotiations,\\ older drivers had a significantly \\ smaller proportion of visual \\ sampling to the left and right.\end{tabular}                        \hspace{50pt}                                                                                    \\
\addlinespace
\begin{tabular}[c]{@{}l@{}} \citet{dukic2012older}\end{tabular} & \begin{tabular}[c]{@{}l@{}}T-USI:right turn,\\ SI: right turn,\\ and left turn\end{tabular}                                & \begin{tabular}[c]{@{}l@{}}Real driving \\ (Eye Tracker)\end{tabular}                         & \begin{tabular}[c]{@{}l@{}}Fixation (each gaze \\ point was associated \\ with one location \\ and one object) of \\ older and youger \\ drivers\end{tabular} & \begin{tabular}[c]{@{}l@{}}Older driver follow same \\ behavioural pattern as \\ younger. Difference found in \\ gaze zone where older driver looked \\ more on markings whereas younger \\ driver looked at dynamic objects.\end{tabular}                                                   \\
\addlinespace
\begin{tabular}[c]{@{}l@{}} \citet{romoser2013comparing}\end{tabular}      & \begin{tabular}[c]{@{}l@{}}USI:Right turn,\\ left turn, and \\ going straight\end{tabular}                                & \begin{tabular}[c]{@{}l@{}}Simulator \\ (Eye Tracker)\end{tabular}                    & \begin{tabular}[c]{@{}l@{}}Older versus \\ younger drivers, \\ glance frequency \\ pattern in each \\ gaze zone\end{tabular}                                  & \begin{tabular}[c]{@{}l@{}}Compared to older drivers, younger\\ drivers spent more time looking at \\ the central area while turning left and \\ less time going straight.\end{tabular}                                                                                                      \\
\begin{tabular}[c]{@{}l@{}} \citet{scott2013visual}\end{tabular}        & \begin{tabular}[c]{@{}l@{}}USI:Right \\ turn (gap \\ acceptance) \end{tabular}                                         & \begin{tabular}[c]{@{}l@{}}Simulator \\ (Eye Tracker)\end{tabular}                    & \begin{tabular}[c]{@{}l@{}}Gaze transitions \\ made by \\ different drivers \\ group\end{tabular}                                                             & \begin{tabular}[c]{@{}l@{}}Young experienced drivers, who \\ are at ‘lower risk’ of accident, \\ showed a \\ more even distribution of gaze \\ compared to the ‘at risk’ groups.\end{tabular}                          \\
\begin{tabular}[c]{@{}l@{}}  \citet{lemonnier2015gaze}\end{tabular}   & \begin{tabular}[c]{@{}l@{}}USI: approaching \\intersection \\ (effect of traffic \\ density and road \\ sign)\end{tabular}              & \begin{tabular}[c]{@{}l@{}}Simulator \\ (Eye Tracker)\end{tabular}                    & \begin{tabular}[c]{@{}l@{}}Gaze accumulation \\ and gaze transitions\end{tabular}                                                                             & \begin{tabular}[c]{@{}l@{}}Visual attention to intersecting roads\\ changed with the priority rule and \\ influenced the visual attention associa\\ -ted with the vehicle control sub-tasks.\end{tabular}                                                                                    \\
\begin{tabular}[c]{@{}l@{}} \citet{li2019drivers}\end{tabular}           & \begin{tabular}[c]{@{}l@{}}SI (Green phase) \\ and USI:left turn, \\ right turn, and \\ going straight\end{tabular}       & \begin{tabular}[c]{@{}l@{}}Real driving \\(Remote \\ setup gaze \\ estimation)\end{tabular}     & \begin{tabular}[c]{@{}l@{}}Glance duration, \\ glance frequency \\ and gaze transition \\ probabilities\end{tabular}                                          & \begin{tabular}[c]{@{}l@{}}Visual scanning performance was \\ similar between SI and USI in  \\ through and right turning \\ maneuvers\end{tabular}                                                                                                                                \\
\addlinespace
\begin{tabular}[c]{@{}l@{}} \citet{lemonnier2020drivers}\end{tabular}    & \begin{tabular}[c]{@{}l@{}}USI: approaching \\intersection \\ (Effect of traffic \\ density, road sign \\ and familiarity)\end{tabular} & \begin{tabular}[c]{@{}l@{}}Real driving \\ (Eye Tracker)\end{tabular}                         & \begin{tabular}[c]{@{}l@{}}Driver visual \\ attention, head and \\ gaze horizontal \\ eccentricity\end{tabular}                                               & \begin{tabular}[c]{@{}l@{}}The effect of expected traffic density, \\ priority rule, and familiarity increases \\ as the distance of the driver to the \\ intersection decreases and temporal \\ pressure increases.\end{tabular}                                                            \\
\begin{tabular}[c]{@{}l@{}} \citet{savage2020effects}\end{tabular}       & \begin{tabular}[c]{@{}l@{}}USI: approaching \\intersection \end{tabular}                                                                                                       & \begin{tabular}[c]{@{}l@{}}Simulator \\ (Eye Tracker)\end{tabular}                    & \begin{tabular}[c]{@{}l@{}}Effect of guidance\\ type (lead vehicle\\ and GPS) and age \\ on eye, head and  \\ eye plus head scan\end{tabular}                 & \begin{tabular}[c]{@{}l@{}}Older driver had lower eye and head \\ scan  compared to younger\\ drivers.\end{tabular}                                                                                                                                                                 \\
\begin{tabular}[c]{@{}l@{}} \citet{chung2022static}\end{tabular}        & \begin{tabular}[c]{@{}l@{}}SI: approaching \\ and performing \\ maneuvers\end{tabular}                                    & \begin{tabular}[c]{@{}l@{}}Simulator \\ (VR driving\\ simulator )\end{tabular}        & \begin{tabular}[c]{@{}l@{}}Static (Dwell time)\\ and dynamic (Gaze \\ transitions) analysis \\ of driver gaze\end{tabular}                                    & \begin{tabular}[c]{@{}l@{}}Novice driver showed longer dwell \\ time and longer fixation duration on\\  AOI.  Gaze transition of novice \\ drivers between AOI occur at close \\distance, while experienced driver \\check surrounding traffic conditions \\ for vehicle driving.\end{tabular} \\
                          \\
\begin{tabular}[c]{@{}l@{}} \citet{lyu2022visual}\end{tabular}        & \begin{tabular}[c]{@{}l@{}}USI: approaching \\ and performing \\ maneuvers\end{tabular}                                    & \begin{tabular}[c]{@{}l@{}}Simulator\\(Eye Tracker)\end{tabular}        & \begin{tabular}[c]{@{}l@{}}Fixation duration,\\number of fixation,\\ gaze transition, \\ blink duration, \\ heart beats rate\end{tabular}                                    & \begin{tabular}[c]{@{}l@{}}Both intersection types and priority \\ rule made differences in drivers'\\scanning behavior and mental \\workloads\end{tabular} 
\\
\begin{tabular}[c]{@{}l@{}} \citet{pike2023gaze}\end{tabular}        & \begin{tabular}[c]{@{}l@{}}USI: approaching \\ (Effect of road \\signs and cross \\walk treatments) \\ \end{tabular}                                    & \begin{tabular}[c]{@{}l@{}}Real driving \\ (Eye Tracker)\end{tabular}        & \begin{tabular}[c]{@{}l@{}}Gaze,\\ gaze transition\end{tabular}                                    & \begin{tabular}[c]{@{}l@{}} No significant effect of road cross-\\walk signs, significantly higher gaze \\value for treatment compared to no \\treatment crosswalk\end{tabular} \\
\\
\begin{tabular}[c]{@{}l@{}} \citet{ringhand2022approaching}\end{tabular}        & \begin{tabular}[c]{@{}l@{}}USI: approaching \\ (Effect of surroun-\\ding traffic, \\intersection type \\driving maneuver, \\secondary task \\involvement) \\ \end{tabular}                                    & \begin{tabular}[c]{@{}l@{}}Simulator \\ (Eye Tracker)\end{tabular}        & \begin{tabular}[c]{@{}l@{}}Relative frequency,\\ of fixation\end{tabular}                                    & \begin{tabular}[c]{@{}l@{}}  Effect of adjacent traffic, secondary \\task involvement and the intended \\traffic changed significantly depen-\\ding on the right of way of the \\driver.\end{tabular}
\\
\addlinespace
\begin{tabular}[c]{@{}l@{}}  \citet{girgis2023drivers}\end{tabular}        & \begin{tabular}[c]{@{}l@{}}SI: Right turn \\ (Left turn in Indian \\driving), safety of \\vulnerable road \\user (pedestrian \\and cyclist) at turns \\ \end{tabular}                                    & \begin{tabular}[c]{@{}l@{}}Real driving \\ (Eye Tracker)\end{tabular}        & \begin{tabular}[c]{@{}l@{}}Glance duration,\\ glance frequency\end{tabular}                                    & \begin{tabular}[c]{@{}l@{}} Glances were more towards the \\pedestrian, rather than cyclist. \\Driver attention was more towards\\ the leftward (rightward Indian driving)\\ traffic at red light of the driver.\end{tabular}\\\hline
\end{tabular}
\label{table:4}
\end{table}

\subsubsection{Over taking/lane changing gaze behavior}
Overtaking or lane-changing occurs in the traffic stream when all vehicles do not move at the design speed \citep{chandra2012overtaking}. In this study, we consider lane-changing gaze behavior for the overtaking of vehicles. A leading vehicle moving at a slow speed hinders the following fast-moving vehicle, provoking the following vehicles to overtake, and lane changing occurs. A sequence of different glance patterns can be seen before the lane change starts to know the possible threats from the surrounding traffic. In lane changing gaze behavior analysis, two kinds of study have been done: driver gaze behavior understanding during a lane change and predicting lane changes based on driver gaze. A detailed discussion of driver gaze-based lane change prediction system is given in Section 6.2.1.

\par In a simulation study, \citet{salvucci2002time} found that drivers began to exhibit notably different gaze behavior about three seconds before the lane change (independent of the vehicle speed), with an increase in the frequency of glances in the rearview mirror at the expense of glances in the direction of their current lane. As soon as the driver decides to change lane, their eyes typically move from salient guiding features of the present lane (such as the tangent point or the lead car) to salient guiding features of the destination lane. Additionally, drivers increase their gazes at surrounding vehicles during lane changes to help with situation awareness and decision-making. Another simulation-based study by \citet{lavalliere2011changing} tested the influence of age differences in glances to the blind spot and mirrors when changing lanes. Compared to younger drivers, older drivers were found to show reduced glance frequency while checking towards the left side mirror, rearview mirror, and blind spot. This behavior of drivers may explain the observations made by \citet{stutts2009identifying} that older drivers are more likely to be involved in collisions when changing lane.

\subsubsection{Driver gaze behavior on curve, on-ramp and off-ramp sections}
This section will cover driver gaze behavior while going through curves, on-ramps, and off-ramps. A study by \citet{lehtonen2014effect} examined the look ahead fixation driving experience behavior when approaching and negotiating the curve on a rural road. They found that experienced drivers spent more time on look-ahead fixation than the road ahead while moving through the curve. To obtain accurate foveal information from the rest of the curve, drivers need to make an eccentric fixation towards the road further up, disengaging the gaze from the visual guidance of online control of steering; these fixations have been called look-ahead fixations. Compared to the entry phase, the driver look-ahead fixation behavior is more on the approach phase because the turning steering driver needs a higher visual demand at the curve \citep{tsimhoni2001visual, lehtonen2014effect}. \citet{zwahlen2003viewing} investigated the effect of ground-mounted diagrammatic guide signs on driver eye scanning before entrance of freeway ramps. The diagrammatic guide sign is a type of sign that indicates the destination using large map-like figures of the road layout. The findings of this study revealed that ground-mounted signs on multi-lane arterials do not excessively distract drivers or influence eye-scanning behavior detrimentally. In situations where placing overhead span type signs can not be economically feasible although placing these signs is highly desirable for driver guidance, these diagrammatic signs give unfamiliar drivers more navigational information in advance (by identifying the correct lane to access the desired entrance ramp).

In addition to real-world experimental studies, simulation studies have also been done to understand driver gaze behavior in on-ramps and off-ramps. \citet{mecheri2022gaze} checked how the presence of pavement shoulder influences driver gaze in the right bend curve on a two-lane rural highway. The results showed that driver gaze shifted towards the inside of the curve, followed by the steering trajectory, irrespective of the width of the shoulder. They suggested that the delineator on the curve is more useful to bring driver gaze and vehicle back to the lane. 
Further, \citet{zahabi2017driver} analyzed the effect of driver age, specific service sign content and format, and familiarity with the road sign on the performance and attention level of the driver when exiting the freeways, i.e., on-ramps and off-ramps. Drivers were found to identify six-panel signs more accurately than nine-panels and were found to be more accurate when familiar with the road sign. 

\subsubsection{Influence of roadside advertising structure on driver gaze behavior}
A recurring finding in the literature is that there appears to be a link between crashes \citep{sisiopiku2015digital, wallace2003driver} and the presence of roadside advertising. In literature, different studies have investigated the effect of different advertisement characteristics on driver behavior. Some of the attributes of the advertisement are its nature, placement, content, road and traffic characteristics, type of area, and driver characteristics \citep{anciaes2022effects}. 
\par Compared to traditional static road signs, electronic roadside advertising has been usually found to have a more significant influence on driver attention, which causes higher safety risk for the general public \citep{beijer2004observed, roberts2013impact, dukic2013effects, herrstedt2017led, oviedo2019impact}. Further, the brightness levels and illumination of roadside advertising billboards also affect visual driver behavior. Driver gaze is attracted when the luminance changes in the visual field \citep{roberts2013impact, oviedo2019impact}. In a simulated study, \citet{herrstedt2017led} examined the impact of LED advertising signs on driver gaze behavior. They found that average glance duration was higher for LED-based signs compared to other objects (e.g., non-Led signs, hoardings, rearview mirrors, speedometer, etc.)  
In comparison to control parts of the road with no billboards, drivers on the section of the road with billboards drove at lower mean speeds, with more speed variability, lane position variability, time spent at high-risk headway, and more visual fixations. The least detrimental effects on driving outcomes were caused by billboards with simple (versus complex) content presented for a longer dwell time (60 seconds versus 40 or 20 seconds). Regardless of dwell time, the billboards with complex content had similar adverse effects on driving. 
\citet{vickers2017animal} observed that the drivers confronting potential risks glanced more at street-level advertisements than the ones raised three meters from the street light. When the vehicle speed is low, drivers have been found to pay more attention to the electronic billboards and other advertising at the junction than other road locations \citep{abbas2020driver}. Further, the frequency of glances has been found to be higher in retail areas \citep{mollu2018driving}, while longer duration of glances has been found in rural areas \citep{costa2019driver}. 

\subsection{Application of driver gaze}
Driver gaze is useful in different applications, such as building driver maneuver recognition and prediction systems, inattention and distraction detection systems, advanced driving assistance systems, etc. In the following section, each of them will be discussed separately.

\subsubsection{Driver maneuver recognition and prediction}
Recognizing and predicting driver intent before the driver's action is important for driver monitoring system. Knowing the driver's intent and correlating it to predict the driver's action is useful to warn the driver for their safety. Driver gaze plays an important role in knowing the driver's intent. 
\par Research on lane change prediction indicates that the driver's glances can be used as an early indicator before a lane change \citep{salvucci2002time, doshi2009roles, pech2014head, martin2018dynamics}. The sequence of lane change on a highway has been defined by considering three parameters: reference point, start time, and end time. The lane marking has been defined as the reference point, while the start point is considered to be the time just before touching the lane marking. Similarly, end time is taken as the time when the tyre just crosses the reference lane marking. Since three seconds is usually taken as the critical decision-making time for a lane change \citep{mourant1974mirror}, several studies considered 3-5 seconds duration before the start time to analyze the driver gaze behavior \citep{salvucci2002time, doshi2009roles, martin2018dynamics, long2022does}. A lane change prediction system helps drivers safely change lanes during overtaking. Lane change prediction based on eye movements in a simulation-based study  proposed a 4DDTW (four-dimensional dynamic time warping) KNN-based lane-changing prediction approach \citep{long2022does}. They used a sliding-space time algorithm to extract the scanpath of the left and right eye in time series data. 4DDTW was used to find the similarity between the scanpaths and then applied KNN on each sample to classify left lane change, right lane change, and lane keeping. The classification accuracy of the system using KNN classifier was 86.5 \%.
In another study, \cite{martin2018dynamics} developed a machine vision-based predictor for lane change behavior. They categorized dynamic lane changes into three classes: left lane change, right lane change, and lane keeping. The analysis focused on a 10-second time window, with 5 seconds before and after crossing the lane marking. Multiple scan paths were examined for each class in each time window, extracting features like minimum, maximum, and average glance duration, frequency, and gaze accumulation in specific zones. The lane changes gaze predictor was modelled using a multivariate normal distribution (MVN), achieving approximately 75.0 \% accuracy in predicting right and left lane changes.
\par {\color{black}\citet{jain2015car} developed an Autoregressive input-output hidden Markov model (AIO-HMM) to predict vehicle maneuvers. This model incorporates a diverse range of input features, encompassing both inside and outside vehicle information. Inside vehicle information, such as driver head movements (facilitated by face detection, tracking, and feature extraction) were combined with outside vehicle information, including road conditions, vehicle dynamics, and GPS coordinates.
The road-facing camera was used to extract binary lane features (left lane, right lane) while integrating GPS coordinates with street maps, which enables the identification of road artifacts within a 15-meter range. Notably, the model encodes the vehicle's average, minimum, and maximum speed over the previous five seconds as additional features. Finally, the model outputs probabilities for four distinct maneuvers (left/right lane change, left/right turn) and driving straight. 
\citet{rahman2020predicting} used a similar AIO-HMM model with  different input features to forecast vehicle maneuvers at intersections. The study integrated five key features, namely 3D driver gaze position, driver head position, traffic light state, steering wheel position, and speed, into a unified feature vector for maneuver prediction. The model was able to anticipate maneuvers (left turn, right turn, going straight) 3.97 seconds before the actual maneuver occurred, achieving an average precision and recall of 86.4\% and 87.7\% respectively.}
 In a simulation study \citet{wu2019gaze} proposed a driver intention prediction system with the driver seated in a semi-autonomous (ego vehicle) vehicle. They proposed a probabilistic Dynamic Time Warping-Hidden Markov Model (pDTW-HMM) to predict the intention of the driver 3 seconds before the execution using fixation and saccades. 

\subsubsection{Driver inattention and distraction detection system}
Several studies show driver distraction and inattention are the leading cause of vehicle crashes and incidents. {\color{black} In a review on driver distraction and inattention, \citet{regan2011driver} defined driver inattention as} ``insufficient or no attention to activities critical for safe driving" and driver diverted attention (which is a synonym of driver distraction) is an alternate form of driver inattention. For a more in-depth understanding, the reader can refer to the studies \citep{regan2011driver, koay2022detecting, kotseruba2022attention} related to driver distraction and inattention. In this section, we will first discuss inattention-based studies followed by driver distraction-based studies.
\par Driver inattention is generally used in literature than attention due to its association with driver safety risk. In two different studies, \citet{fletcher2009driver} and \citet{hu2021data} developed a driver inattention detection system using eye gaze and road events. Multiple road event inattention detection systems were built, and one of them was road center inattention detection system. A warning was given whenever the driver can be seen to be diverted from the forward direction for a minimum specific time period. 
\par One of the important measures of driver distraction is the measure of eyes off the road \citep{vicente2015driver, badgujar2023driver}. Eyes off the road indicates that the driver is looking somewhere else rather than looking at the road. This situation delays the reaction time of the driver and causes a higher risk of accidents \citep{lamble1999detection, bargman2015does, dingus2016driver}.  {\color{black}\citet{ahlstrom2013gaze}  developed an attention-based driver warning system called AttenD that can help to mitigate driver distraction. This study assumes that the driver attention is directed towards the gaze object. A 3D car model was divided into several zones, such as mirrors, speedometer, etc. 
The algorithm works according to the principle that frequent glances away from Field Relevant for Driving (FRD) can be used as a sign of distraction. FRD is defined as the intersection between a viewing cone of 90 degrees and vehicle windows for gaze tracking. The authors considered 2 seconds as the buffer time, which will depleted when the driver gazes away from the FRD. When the buffer time reaches zero, a warning is given to the driver.}

 \par Apart from driver distraction systems, studies have also been performed to measure the cognitive load of drivers. \cite{prabhakar2020cognitive} developed a system to measure driver's cognitive load in both simulated and real driving scenarios. They classified the cognitive state into two classes: No Task (driving without secondary task)  and Task (driving with secondary task). Two physiological parameters, gaze and pupil based metrics captured by head mounted eye tracker were used as input to the machine learning model. The input gaze parameters were saccade rate, fixation rate, and median velocity of saccadic intrusions (SI), while the pupil parameters were the L1 norm of the spectrum (L1NS), standard deviation of pupil (STDP), and low pass filter of pupil (LPF) from pupil dilation. Two binary classifiers, support vector classifier (SVC) and feed-forward neural network (NN) models were used to define the cognitive state of the drivers. This achieved an accuracy of 75.0 \% using feed-forward NN model.  Using the cognitive state of the driver and detecting eyes-off-road distraction, they have also developed a multimodal driver alert system.

\subsubsection{Building advanced driver assistance system}
An advanced driver assistance system (ADAS) is an automated system that assists the driver when the driver fails or misses some events in nearby traffic \citep{fletcher2005correlating, yang2021driver}. 
These systems use sensors, cameras, radar, and other advanced technologies to collect and analyze data in real-time, providing assistance to the driver in various aspects of driving. There are some key features and ways in which ADAS helps drivers. These key features are collision avoidance system, lane-keeping assistance system, adaptive cruise control, blind spot detection, parking assistance, traffic sign recognition, driver fatigue monitoring, etc. In this study, we have discussed the studies that included the driver gaze in building ADAS.
\par Since driver gaze information is an important input variable to measure driver attentiveness towards the surrounding traffic environment, it plays an important role in building an advanced driver assistance system. In real-world driving scenario, \citet{fletcher2005correlating} made a driver assistance system using driver gaze direction and speed limit sign. When a traffic sign is recognized, the system checks two things: (a) whether the driver looking at the sign and  (b) if the speed and acceleration of the vehicle are compliant with the sign. If the vehicle state is not compliant and the driver has not seen the sign, a high-priority warning is given to the driver. A study by \citet{prabhakar2020interactive} developed an intelligent interactive head-up (HUD) display on the windscreen that does not require taking eyes off the road while doing secondary tasks such as watching navigation map, operating music player, or operating vent control. The system consists of a screen in which content is displayed on the windscreen of the car, and the content on the screen either changes by eye gaze or finger movement of drivers.
\par ADAS technologies collectively contribute to improved safety, reduced accidents, and enhanced driving comfort. However, it is important for drivers to remain attentive and not entirely rely on these systems, as they are designed to assist rather than replacing human judgment and control. Additionally, the effectiveness of ADAS may vary depending on factors such as system quality, environmental conditions, and proper maintenance.

\section{General Discussion and Future Scope}
Over the past decade, significant progress has been made towards driver gaze estimation by gaze zone classification and tracking. Driver gaze data collection methods and equipment have also evolved over the years. Gaze estimation has evolved from coarse gaze zone classification to finer gaze zone classification, and gaze classification models have been developed using traditional machine learning to deep learning algorithms. Detected driver gaze is used to analyze the awareness and attentiveness of the drivers to the surroundings, build advanced driving safety systems, and develop safer driving guidelines for the drivers. However, there are still research gaps existing in the current studies that can be handled in the future, thereby leading to further evolution of this domain and helping to build a safer transportation system.

\subsection{Challenges in driver gaze estimation and possible solutions}
Driver gaze estimation is a challenging task due to the complex and dynamic nature of driving. 
Some of these challenges include varying lighting conditions, occlusions and eyeglasses, dynamic road and traffic environments, in-car distraction, real-time processing, safety concerns, user comfort and acceptance, etc. There are some representative methods that aim to address these challenges in driver gaze estimation. In this subsection, we discuss the limitations and future scope of driver gaze estimation in three broad categories: (a) inclusion of advanced sensing technologies, (b) generating datasets that can better represent real-world conditions, and (c) advancements of models for driver gaze estimation. Each of these is discussed in detail next.

\subsubsection{Sensors}
The datasets mentioned in Table \ref{table:1} typically captured the driver face using a single camera. However, in case of large head movement, one side of the face maybe be occluded by the other, thereby creating problems in detecting the iris or pupil position inside the eye. This in turn affects the estimation of the point of gaze of the driver. These limitations can be overcome using multiple cameras and capturing the driver face at different angles or positions \citep{yang2019dual, wang2022dual}.  Varying light conditions, either due to daytime or due to weather conditions, which impact the gaze estimation accuracies, can be resolved by using NIR or  depth camera \citep{naqvi2018deep}. 

Also, driver gaze can be estimated in terms of object of interest e.g. surrounding vehicles, pedestrians etc. This approach is different from the traditional approach of estimating the gaze zone class or the point of gaze. To estimate the object of interest of the driver, the scene camera of eye tracker or dashboard camera has been used. However, it can be extended further by including information from cameras placed at the top of the vehicles or other sensors such as LiDAR (Light Detection and Ranging) or radar. This can help in accurate estimation of objects of interest of drivers and correlate them with driver actions. Also, monitoring additional cues such as hand movements, facial expressions, or physiological signals can help to differentiate between intentional gaze shifts from distractions \citep{wang2010real, koay2022detecting}.

\subsubsection{Dataset generation} 
As discussed in Section 4, benchmark open-source driver gaze data, discussed in Table \ref{table:1}, have been typically collected in parked or moving vehicles. Mostly, the parked vehicle participants have been college or university students \citep{ribeiro2019driver, dua2020dgaze, ghosh2021speak2label}, who can have limited knowledge of how head movement or eye movements occur during real driving scenarios. On the other hand, datasets collected in real driving scenarios have been limited in terms of the number of participants (drivers) 
 \citep{nuevo2010rsmat, diaz2016reduced, Jain_2015_ICCV, schwarz2017driveahead, palazzi2018predicting}. A limited number of participants in real-world driving datasets therefore lack diversification in terms of experience, age, lighting conditions, and traffic conditions. Therefore, large scale open source driver gaze datasets can be created encompassing different environmental and lighting conditions, which can help in the development of a robust and generalized system of gaze estimation based on deep learning techniques.  

\par Further, there are inherent problems in classifying gaze zones inside a parked vehicle compared to real-world driving scenarios. Driver gaze collection and ground truth generation can be easier inside the parked vehicle and is also safer for the driver. However, this method may affect the psychological behavior of the drivers because, in this method, a second person is providing instructions \citep{costa2019driver, ghosh2021speak2label} to the driver to look towards a particular gaze zone or the driver has been informed earlier to look at predefined gaze zone.
 \cite{chuang2014estimating} revealed that the gaze zone classifier created using parked vehicles could not successfully generalize to a moving vehicle. On the other hand, while doing driver gaze data collection in moving vehicles, drivers gaze naturally in and around the dashboard and windshield area, but generating the ground truth labels of gaze zone is difficult in this situation. Since, in this method, drivers can not look towards the gaze zone by instructions due to accident risk, unsupervised techniques can be used for estimating the gaze zone class \citep{chuang2014estimating}.

\par Apart from data collection strategies, there are also limitations in ground truth generation methodology for gaze datasets. Datasets collected in the parked vehicle have been typically annotated by two or more human annotators and cross-verified. However, these datasets do not typically mention any data statistics on how much the difference or error was observed during ground truth generation of the different annotators and how to handle such cases. Also, gaze data annotations by humans cannot be assumed due to 100\% accurate, which can impact gaze classification algorithms too. 
Ground truth generation by speak2label \citep{ghosh2021speak2label} is an automatic way of ground truth generation, but it still has the limitations of data generation inside parked vehicles, as discussed above. 
Ground truth labels of existing driver gaze datasets are typically given based on the zone in which the driver is looking. These labels are the predefined regions such as rearview mirror, forward, left wing mirror, right wing mirror, center stack, speedometer, etc. These datasets do not clearly mention the surrounding data collection conditions except DGAZE. During recording the DGAZE datasets, moving traffic was shown on the screen, and the gaze zones were based on the vehicle entities \citep{dua2020dgaze}. Therefore, in future, gaze datasets can be classified not only based on the gaze zone but can also denote the vehicle entities observed, which can help in understanding driver gaze behavior better.

\subsubsection{Models for gaze estimation}
\par Driver gaze zone estimation using conventional appearance-based method and appearance based method with deep learning has certain limitations. In conventional appearance based method, the decision made by the traditional ML model completely depends on the individual sub-model (face, pupil detection, landmark estimation, feature extraction), which affects the accuracy of the gaze estimation. The hand-crafted features designed from facial landmarks on the eyes are not completely robust to variations across different drivers, cars, seat positions, etc.
On the other hand, gaze estimation using appearance based method with deep learning, uses pre-trained CNN model which does not require hand-crafted features because of inherent feature extraction quality. However, these models require large-scale datasets and more computational power for their training compared to traditional ML models. The availability of large-scale open-source datasets with different illumination conditions and more subjects can help build a robust classifier. 
On the other hand, geometric model-based methods suffer in remote setup-based driver gaze estimation due to varying light conditions and the difficulty of capturing geometric eye parameters in non-constrained environments. However, in head-mounted setup based gaze estimation, this method can be used to give more accurate results compared to appearance-based methods. However, in both these methods, attention can be given to developing calibration-free gaze estimation techniques, which can help in large-scale implementation purposes \citep{tonsen2020high}.

\par Gaze estimation models can also focus on estimating the exact point of gaze instead of gaze zones in the windscreen, right and left-wing mirror only. This can help to understand driver attention on different traffic entities, such as vehicles, pedestrians, etc., and other surrounding objects, such as billboards, traffic signs, etc. This can also help to understand driver behavior during complex traffic manoeuvres such as intersections, on-ramps, off-ramps, etc.

\par Finally, driver gaze estimation models can be further advanced based on the techniques that have been successful in gaze estimation in other platforms, such as logistics and manufacturing, gaming, virtual reality, and human-computer interaction (HCI) systems.   For example, gaze estimation based on attention mechanisms has been an active area of research in the field of computer vision and human-computer interaction. Recently, several studies based on attention mechanisms developed human gaze estimation systems in different fields \citep{murthy2021appearance, fang2021dual, senarath2022customer}. These studies have mostly focused on developing depth-based dual attention model for customer gaze estimation in retail stores \citep{senarath2022customer} or human gaze estimation in the wild \citep{fang2021dual}. Similar attention-based mechanism can be developed in future for 3D driver gaze estimation in wild.

\subsection{Future scope in uses and applications of driver gaze}
\par There has been a significant number of studies on driver gaze behavior that have focused on maneuvering through intersections, lane changing during overpassing, etc. However, most of these studies are either based on comparing different driver groups, such as younger versus older, or based on experience, such as novice versus experienced. A few studies, based on real driving \citep{lemonnier2020drivers} and simulator \citep{lemonnier2015gaze}, evaluated the impact of traffic density and the familiarity of the route at intersections on the driver gaze. Further, gaze behavior studies on mixed traffic conditions and unstructured driving environments are limited; only one study in real driving compared the gaze pattern of drivers on SI and USI \citep{li2019drivers}. Hence, more research is needed on driver gaze behavior in mixed-traffic environments, which can include the effect of pedestrians, traffic density, intersection type along with focusing on unstructured driving environments too.

\par Eye tracker based driver gaze behavior studies use different terminologies such as fixation, dwell time, and saccades; on the other hand, studies based on the remote setup use glance duration, glance frequency, and gaze transitions to define driver gaze behavior. They can explain the gaze behavior based on statistics such as minimum, and maximum glance duration, glance frequency, number of fixations, dwell time in each gaze zone, glance or gaze transitions, saccades between the gaze zones, etc. However, these studies do not consider the effect of the shape and size of different traffic entities on driver gaze. For example, is the influence of small vehicles (cycles, motorcycles) on driver gaze same or different than the large vehicles such as trucks, buses, etc.? More in-depth analysis on understanding such gaze behavior will be helpful in determining driver attentiveness \citep{yamaguchi2022estimation} and building a safer driving environment.

\par Driver gaze research can also be incorporated and extended with other physiological sensors \citep{wang2010real} such as heartbeat (fitbit) \citep{ lu2022detecting, othman2022drivermvt}, EEG (Electroencephalography) \citep{tuncer2021eeg, arefnezhad2022driver} to understand driver gaze behavior comprehensively. Monitoring the pulse rate, breathing rate, and other physiological data might reveal the psychological state of the driver \citep{abou2020application, zhou2023driver}. Since connected and autonomous vehicles are the emerging future of intelligent transportation systems, more attention is required to know how the driver gaze is influenced by surrounding traffic, including autonomous or connected vehicles. Limited research has been done on the use of driver gaze in building driver assistance systems. So there is significant scope for researchers and industry experts to build a robust and generalized driver assistance system that can help drivers perform safe maneuvers at intersections, overpassing, etc. Driver gaze behavior understanding can also help in taking steering control from manual to semi-automatic when the driver is not fully attentive to the surrounding traffic environment. 

\section{Conclusion}
Driver gaze plays an important role in different driving gaze-based applications, such as driver inattention detection, visual distraction detection, and taking over automatic steering control.
This study thoroughly summarizes different terminologies used in driver gaze estimation and behavior understanding based on head-mounted and remote setup-based techniques and compiles the existing benchmark gaze estimation datasets. We have also reviewed different algorithms and models used in driver gaze estimation based on appearance and geometric model approaches. Compared to traditional machine learning, deep learning-based approaches are more efficient in detecting driver gaze in different lighting conditions and large head movements. 
This study also explores uses of driver gaze in different driver gaze behavior understanding such as negotiating at intersections, lane changing during overpassing, moving on the curve, on-ramp and off-ramp, influence of roadside advertising infrastructure and driver gaze-based applications such as driver maneuver recognition and prediction, inattention and distraction detection, and ADAS. Finally, we have highlighted the limitations of the existing studies and the future scope in this domain, which can help the researchers and the developers to build a more robust and generalized driver gaze estimation and gaze-based driving assistance system.

\section*{CRediT authorship contribution statement}
\textbf{Pavan Kumar Sharma:} Conceptualization, Formal analysis, Investigation, Methodology, Writing-Original Draft.
\textbf{Pranamesh Chakraborty:} Conceptualization, Funding acquisition, Investigation, Methodology, Supervision, Writing-Review \& Editing.

\section*{Declaration of competing interest}
The authors declare that they have no known competing financial interests or personal relationships that could have appeared to influence the work reported in this paper.

\section*{Data availability}
No data was used for the research described in the article.

\section*{Acknowledgement}
Our research results are based upon work supported by the Initiation Grant scheme of Indian Institute of Technology Kanpur (IITK/CE/2019378). Any opinions, findings, and conclusions or recommendations expressed in this material are those of the author(s) and do not necessarily reflect the views of the IITK.

\bibliography{bibliography}
\end{document}